\documentclass[sigconf,nonacm]{acmart}

\AtBeginDocument{%
  }

\setcopyright{acmlicensed}
\copyrightyear{2026}
\acmYear{2026}
\acmDOI{XXXXXXX.XXXXXXX}

\usepackage{soul}
\usepackage{url}
\usepackage{graphicx}
\usepackage{amsmath}
\usepackage{algorithm}
\usepackage{comment}
\usepackage{algpseudocode}

\usepackage{tikz}
\usepackage{multicol}
\usepackage{colortbl}
\usepackage[svgnames]{xcolor}

\usepackage{enumitem}

\usepackage[most]{tcolorbox}
\usepackage{listings}
\usepackage{tikz}
\usetikzlibrary{arrows.meta,positioning}
\usepackage{placeins}
\usepackage{multirow}
\usepackage{pifont}

\usepackage{tikz}
\usepackage{subcaption}
\usetikzlibrary{shapes.geometric, arrows}

\tikzstyle{startstop} = [rectangle, rounded corners=0.5cm, minimum width=3cm, minimum height=1cm,
text centered, 
fill=red!20]

\tikzstyle{io} = [rectangle, rounded corners,
minimum width=3cm, 
minimum height=1.5cm, 
text centered, 
text width=2.5cm, 
fill=blue!30]

\tikzstyle{io2} = [rectangle, rounded corners,
minimum width=3cm, 
minimum height=1.5cm, 
text centered, 
text width=2.5cm, 
fill=yellow!30]

\tikzstyle{process} = [rectangle, rounded corners,
minimum width=3cm, 
minimum height=1.5cm, 
text centered, 
text width=2.5cm, 
fill=orange!30]

\tikzstyle{decision} = [rectangle, rounded corners, 
minimum width=3cm, 
minimum height=1.5cm, 
text centered, 
text width=2.5cm, 
fill=teal!30]
\tikzstyle{arrow} = [thick,->,>=stealth]
\begin{document}
\begin{sloppypar}
    
\title{Geo-Spatial Concept Probing of Large Language Models:  \\ Abstraction, Compositionality, and Grounding }

\author{Karim Radouane}
\email{karim.radouane@irit.fr }
\affiliation{%
  \institution{University of Toulouse, IRIT}
  \city{Toulouse}
  \country{France}
}

\author{Jose G Moreno}
\email{jose.moreno@irit.fr }
\affiliation{%
  \institution{University of Toulouse, IRIT}
  \city{Toulouse}
  \country{France}
}

\author{Lynda Tamine}
\email{lynda.tamine@irit.fr }
\affiliation{%
  \institution{University of Toulouse, IRIT}
  \city{Toulouse}
  \country{France}
}

\begin{abstract} 
Understanding concepts is fundamental to generalization. Despite their impressive performance on a wide range of tasks, Large Language Models (LLMs) still struggle with genuine concept understanding. Prior work has evaluated conceptual understanding in LLMs using natural-language benchmarks or narrowly scoped synthetic tasks, but these settings often conflate multiple skills or lack precise control over the underlying concepts and their properties. To support controlled probing of concepts in LLMs, we design tests on their core properties: abstraction, compositionality, and groundness. We set up a concept-centric benchmark,  targeting spatial concepts such as direction, distance, topology, and their compositions, and use question answering tasks serving as a proxy. We conduct extensive experiments across multiple LLM architectures and training regimes to analyze how model scale and design impact conceptual understanding. 
The results reveal clear limitations in current LLMs and provide insights into the factors shaping their ability to acquire and compose structured concepts.  Our findings shed light on how concept-based LLMs can be redesigned for improved information access and knowledge management. The code will be available at \url{https://github.com/rd20karim/concept-probing}.

\end{abstract}



\maketitle
\section{Introduction}\label{intro}
Large Language Models (LLMs) exhibit a wide range of capabilities, but the question of whether they can be models of human language understanding is still an open philosophical and scientific debate that impacts cognitive science, machine learning (ML), and natural language processing (NLP), and information research disciplines \cite{Pavlick23,YILDIRIM2024404,Goddu24}, to cite but a few. One major subject of debate is the source of their capabilities: do they behave as \textit{statistical parrots}, or are they able to organize symbolic, structured representations of \textit{concepts} that they manipulate in the generation process? The question is key since concepts are the cornerstone of intelligence, allowing better generalization,  and interpretability \cite{Poeta25}. \\
Recent work has developed probing and learning techniques for concepts in LLMs. Probing can focus on factual knowledge through downstream task evaluation \cite{manvi2024geollm,zhang2025geoanalystbench,ji2025foundation,Dumitru25} or on their internal representations to explain their predictions (e.g., \textit{mechanistic interpretability} \cite{wang-etal-2024-knowledge-mechanisms,Parry25}). Concept learning refers to the learning of concept representations and their explicit utilization in the internal layers of models, such as \textit{concept bottleneck models} \cite{koh2020concept}. \\
Given the open-ended nature of concepts across tasks and domains, other works evaluated whether \textit{concept properties} acknowledged in the literature \cite{Fodor1975-FODTLO,Stein24,lovering-pavlick-2022-unit,lewis-etal-2024-clip} (e.g., compositionality \cite{Stein24,lewis-etal-2024-clip}),  can be represented or even learned  in LLMs. Evaluation of \textit{concept property} provides a more fine-grained insight into the \textit{concept understanding} of LLMs, enabling better control over both concept probing and learning.  \\

\noindent\textbf{Research gap.}
Recently, there has been growing interest in studying LLMs' understanding of concepts such as truth \cite{azaria-mitchell-2023-internal}, time and space \cite{gurnee2024language}, patient gender \cite{ahsan-etal-2025-elucidating}, and gender bias \cite{yu2025understanding}. Other works addressed the study of concept properties such as  complexity \cite{jin-etal-2025-exploring},  compositionality \cite{Stein24,lewis-etal-2024-clip}, and grounding  \cite{park2025iclr,li2023can}.
However, while previous work provides insights on the abilities of LLMs to succeed in (cross-modal) tasks through task performance involving concepts of interest, they do not systematically assess concepts per se for the following reasons: (i) they mostly consider multi-modal LLMs (text and image) and concept probing is addressed through representation sharing between modalities (e.g., \cite{li2023can}); (ii) the few works that considered text-only LLMs covered only one core property of concepts (e.g., abstraction \cite{gurnee2024language}, grounding \cite{Pavlick23}) limiting the scope of concept probing;  (iii) the design of probes is not guided by the concepts per se, but instead by downstream tasks, thereby confounding conclusions about both LLMs' concept understanding and task skills (e.g., \cite{Ramrakhiyaniipm2024,yamada2024evaluating}). \\ 

\noindent\textbf{Goal and research questions.} In this work, we seek to fill the above research gap by investigating whether LLMs can build, from only the \textbf{text modality}, \textbf{concept} representations of world models using a \textbf{controlled concept-centric benchmark}.  Inspired by previous work \cite{lovering-pavlick-2022-unit}, we approach this challenge by framing it as a set of \textbf{probing tests} about \textbf{each of the core properties} of conceptual representations: \textbf{abstraction}, \textbf{compositionality}, and \textbf{grounding}. Under this perspective, our main goal can be formalized as:\\
\textit{How can we translate \textit{abstraction}, \textit{compositionality}, and \textit{grounding} into testable hypotheses and develop the empirical probing tests allowing us to quantify the extent to which LLMs align with? }

We choose three popular spatial concepts---\textit{direction}, \textit{distance}, and \textit{topology}---which are particularly challenging for dealing with spatial commonsense since they are rarely expressed in texts, though our probing methodology is designed to generalize across spatial concepts and extend to other concepts, tasks, and domains. 
Specifically, we address the following research questions:

\noindent(RQ1) Are LLMs’ performances on a geo-spatial question-answering task sufficient to assess their actual ability to understand underlying concepts?

\noindent(RQ2) Do LLMs encode abstract geo-spatial concept representations? Are concept representations generalizable across linguistic tokens and geographic regions?\\
(RQ3) Are LLMs able to compose geo-spatial concepts? Can LLMs' predictions be explained by composing concept representations?\\
(RQ4) Can concept representations be grounded in real-world knowledge? \\

We build a \textbf{concept-centric benchmark} and probe popular text-only LLMs (Llama-8B, Mistral-7B/8B, Qwen-0.6B, 1.7B, 4B, 8B)  through binary and multi-choice question-answering (MCQ) tasks to analyze their internal representations across layers and concepts.
\noindent\paragraph{\textbf{Main findings}}
\begin{itemize}

    \item For RQ1, we assess LLM performance on our benchmark as a preliminary proxy for concept understanding. We find moderate MCQ accuracy  coexists with low consistency under Binary-QA, exposing unstable concept understanding.
    This limitation is most analytically tractable for the \textit{distance} concept. 
    \textbf{These findings clearly motivate moving beyond geo-spatial QA performance toward targeted probing of core conceptual properties.}
    
    \item For RQ2, we measure the extent to which concepts are explicitly encoded in LLMs, and their instances represent types as an abstraction across tokens and regions. Our findings indicate that most models encode concepts with a good level of generalization across instances, except the Mistral family models wich consistently underperform across generalization splits and layers. 
    \item For RQ3, compositionality is assessed through: the \emph{compositionality gap}, the additivity of latent representations, and the extent to which predictions can be explained by the composition of latent embeddings and/or atomic predictions. We find that most LLM variants exhibit a compositional structure from a cosine similarity perspective, except for the Mistral models. The correlation analyses using different proposed approaches reveal that compositionality is a key factor for successful concept understanding. In particular, models that preserve correlation across layers perform well, whereas the Mistral family models fail to meet these criteria.
    \item For RQ4, despite enriching concept-based questions with real-world factual information, such as geometrical measurements, it does not improve related QA-task performance in LLMs even with an explicit distance threshold definition. This limitation highlights a persistent gap in the LLMs models’ ability to ground geo-spatial concepts, revealing a misalignment between representations in linguistic space and their corresponding numerical representations.

\end{itemize}


\section{Related Work}\label{state-art}

The primary objective of this work is to investigate LLMs' understanding of concepts through abstraction, compositionality, and grounding. As a use case, we consider geography-related concepts and use performance on question-answering and probing tasks as a proxy. Therefore, we review previous work on (i) concepts in LLMs; (ii) probing LLMs on geography knowledge, and (iii) geo-spatial question-answering. Table 1 compares prior close work on probing LLMs' concepts with ours.

\subsection{Concepts in LLMs}
The definition of \textit{concept}, which has recently gained renewed attention in explainable artificial intelligence (XAI), remains elusive and varies across disciplinary boundaries (e.g., cognitive science, ML \cite{schwalbe2022concept,Poeta25,Fodor1975-FODTLO}). 
In \textit{cognitive science}, a \emph{concept} is  a fundamental unit of structured knowledge representing an interpretable abstraction of entities (e.g., shape, color)  that functions like a symbol used for problem-solving, reasoning, or organizing information \cite{Fodor1975-FODTLO, Goguen2005,Xie25}. In \textit{ML}, a concept is  an abstract representation that captures a meaningful class, attribute, or feature within a model’s learned space that correlates with high-level semantic \cite{Stein24,Fong2018Net2VecQA,polley2022xvision}. Across these disciplines, a conceptual representation is generally assumed to require three core properties \cite{Fodor1975-FODTLO,Stein24,lovering-pavlick-2022-unit,lewis-etal-2024-clip}: (i) \textit{abstraction}\footnote{Abstraction is also referred to as \textit{systematicity.} \cite{Fodor1975-FODTLO}}: different instances of a concept evaluate to the same semantic type. 
This property is strongly related to the ability of concept representations to tackle out-of-distribution generalization \cite{lovering-pavlick-2022-unit,geva-etal-2022-transformer}; (ii) \textit{compositionality}: a concept can be \textit{atomic} or composed of a set of concepts. A \textit{composite} concept is a function of its constituents. Previous work addressed the compositionality of LLMs from two sides. \textit{Compositional reasoning} is the ability to process reasoning chains \citep{PressZMSSL23,lake2018generalization,rosen2011discrete,ji2023ccr} while \textit{conjunctive compositionality} is the ability to disentangle and combine the concept constituents (e.g., via logical AND~\cite{naito2021revisiting}) to recall or understand the composite concept \citep{mikolov2013distributed,naito2021revisiting};
and (iii) \textit{grounding}: universally refers to the fact that concepts are assumed to apply to situational context reflecting physical things in the real-world \cite{Fodor1975-FODTLO, li2023can,yamada2024evaluating}. Grounding tests \cite{beinborn-etal-2018-multimodal} generally rely on multimodal mapping between concept modalities, where a ``modality'' could be either a sensor (e.g., text vs. image) or a mode (e.g., plain text and a textual description of a grid \cite{park2025iclr}).

The question of whether LLMs can give rise to complex conceptual representations has led to a large body of work that falls in the mechanistic interpretability \cite{Nainani2024EvaluatingBM,geva-etal-2022-transformer,Parry25}.  It has been shown that concepts such as truth \cite{azaria-mitchell-2023-internal}, time and space \cite{gurnee2024language}, patient gender \cite{ahsan-etal-2025-elucidating}, and gender bias \cite{yu2025understanding} can be localized  in LLM representations. In their recent work \cite{jin-etal-2025-exploring},  Jian et al. categorized the complexity of concepts based on their level of abstraction and showed that tasks requiring complex concepts require deeper layers for an accurate understanding. Other works studied, particularly, whether concept properties such as compositionality \cite{PressZMSSL23,Stein24,lewis-etal-2024-clip} and grounding  \cite{park2025iclr,li2023can,yamada2024evaluating} hold in LLMs through multi-modal downstream tasks (i.e., using texts and images).

In this work, we probe LLMs from a concept-centric perspective that explicitly models concepts and their three core properties—\textit{abstraction, grounding, and compositionality}—and, unlike prior work, introduce a concept-guided probing benchmark to evaluate LLMs’ internal concept representations and the sensitivity of natural language question-answering performance to the presence of these concepts along with their three core properties.

\vspace{-0.2cm}
\subsection{Probing LLMs for geo-spatial knowledge}
Probing LLMs for geo-spatial knowledge offers a lens into two categories of work. The first category aims to evaluate  the capabilities of LLMs to recall geographic facts using  downstream tasks' performance as a proxy \cite{manvi2024geollm,zhang2025geoanalystbench,ji2025foundation,van2025opportunities}.  For instance, \citet{manvi2024geollm}  demonstrated that LLMs are highly sensitive to prompt formats but that fine-tuning using map data from OpenStreetMap enhances the accuracy and robustness of these models. 
However, it has been shown that fine-tuning LLMs on specific datasets introduces bias against geographic areas with lower socio-economic conditions \cite{ManviIcml24}. \\
The second category of work \cite{gurnee2024language,patel2022mapping} aims to examine the association between internal representations and extrinsic properties of geo-spatial knowledge. 
\citet{gurnee2024language}  showed that LLMs internally store concepts like
latitude, longitude, and time in the early layers, and they can predict a location map. The work developed by  \citet{patel2022mapping} represents a notable conceptual advancement in understanding the capabilities of LLMs to ground the learned conceptual representation --built upon a linguistic world-- with real-world representation based on a few in-context examples. They showed that large-size models exhibit high-level capabilities for grounding and generalization, indicating that they can learn a latent structure of the world. 

\vspace{-.25cm}

\subsection{Geospatial question-answering}

Geospatial question answering (GeoQA) aims to enable generation or retrieval models to answer questions involving geographic entities or concepts and that require spatial operations \cite{MaiJ0CL21}.  
\citet{MaiJ0CL21} report many challenges in GeoQA systems, among which are the \textit{vagueness of geographic concepts} invoked in the questions, the \textit{difficulty in identifying the correct spatial relations between the entities} involved in the questions, and the difficulty in encoding spatial entities. 
With the advancement of LLMs, a large body of work examined their capabilities to address these challenges, by covering a wide range of GeoQA tasks, including place retrieval \cite{Mai24}, and spatial reasoning on topological relations \cite{ji2025foundation,dihan2025mapeval,cohn2025evaluating,deng2024k2}. The main limitations of LLMs that arise from this literature are: (i) LLMs are not spatially replicable over the globe, nor efficiently robust across tasks; (ii) LLMs are not efficiently transferable to tasks handling different spatial scales; and 
(iii) LLMs'   do not generalise the “neighbourhood” relationship.\\ 
The challenges mentioned above motivate us to: (i) target spatial concepts as a use case for our probing study, but without loss of generality; and (ii) design a question-answering benchmark test built upon these concepts.


\begin{table}[t]
\centering
\small
\resizebox{\columnwidth}{!}{%
\begin{tabular}{l|c|c|c|c|c|c|c|c|c c}
\hline
\textbf{Related Work} 
& \multicolumn{4}{c|}{\textbf{Concepts}} 
& \multicolumn{3}{c|}{\textbf{Properties}} 
& \multicolumn{2}{c}{\textbf{Probing Tests}} \\
\cline{2-10}
 & \textbf{Dir}  & \textbf{Dis} & \textbf{Top} & \textbf{Other} & \textbf{Abst.} & \textbf{Comp.} & \textbf{Ground} & \textbf{Knowledge}& \textbf{Repr.} \\
\hline
\citet{Ramrakhiyaniipm2024}            
& - & - & - & Location & - & - & - & \ding{51} & - \\
\citet{patel2022mapping}            
& \ding{51} & - & - & Color & - & - & \ding{51} & - & \ding{51}\\
\citet{manvi2024geollm}            
& - & - & - & Location & - & - & - & \ding{51} &  - \\
\citet{gurnee2024language}            
& \ding{51} & - & - & Time & \ding{51} & - & - & \ding{51} & \ding{51}  \\
\citet{park2025iclr}            
& \ding{51} & - & - & - & - & - & \ding{51} & - & \ding{51} \\
\citet{yamada2024evaluating}      
& - & - & - & - & - & - & - & \ding{51} & - \\
\hline
\textbf{Our work} 
& \ding{51} & \ding{51} & \ding{51} 
& - & \ding{51} & \ding{51} & \ding{51} 
& \ding{51} & \ding{51} \\
\hline
\end{tabular}%
}
\caption{Related work on concept probing in LLMs with a focus on text modality. The table summarizes investigated concepts and properties, and categorizes probing tests by whether they target knowledge or/and internal representations. }
\vspace{-0.8 cm}
\end{table}

\section{Background, Terminology and Notations}
\subsection{Definitions}
\paragraph{\textbf{Concept}.} 
Across disciplines, a \textit{concept}  $C$ is a high-level and  human-interpretable unit of information. Specifically, we adopt a definition rooted in knowledge representation \cite{brachman-1979-taxonomy} and widely used in Information Retrieval (IR)  and NLP, where a concept is an abstract category or class that defines a set of object entities sharing common properties and roles (e.g., concept of \textit{direction}). Concepts are interpreted as unary \textit{predicates}, and roles as binary \textit{relations} between object entities.
\noindent\paragraph{\textbf{Concept instance.}}  A concept instance $c$ is a concrete exemplar of a concept, representing a specific real-world entity that belongs to the extension of that concept. In IR and NLP, a concept is approximated by a finite vocabulary of representative lexical words $c$ that represent its \textit{instances} (e.g., \textit{east}, \textit{west} are instances of concept \textit{direction}).  We note $\mathcal{I}(C)$ the finite representative set of instances of concept $C$. For the sake of simplicity, we consider in practice $C=\mathcal{I}(C)$.
\noindent\paragraph{\textbf{Concept instance representation}.} We consider symbolic concept instance representations
using   binary relational predicates  that express \textit{facts}. A fact is represented as a \textit{positive triplet} $\langle x, r, y\rangle$, where $x$ is the subject entity, $r$ is the core relation (predicate) expressing the concept instance $c$, 
and $y$ is the object entity (e.g., $\langle$\textit{Prescot}, \textit{west\_of}, \textit{Todmorden}$\rangle$ represents \textit{west}, an instance of concept \textit{direction} viewed as a the binary relation \textit{west\_of} between \textit{Prescot} and \textit{Todmorden} object entities). We also consider the negated triplet $\langle x, \lnot{r}, y\rangle$ which embeds a fact with an opposite relation to $r$ (e.g., $\lnot$\textit{west\_of} is \textit{east\_of}). 

\paragraph{\textbf{Concept properties}.} Following \cite{Fodor1975-FODTLO,lovering-pavlick-2022-unit}, we consider three core properties of concepts:
\begin{itemize}[itemsep=0pt, topsep=0pt, leftmargin=10pt]
    
\item   \textit{Abstraction:} a concept $C$ is assumed to be the abstraction of  its instances $\mathcal{I}(C)$. Therefore, $\mathcal{I}(C)$ form the \textit{semantic type} of $C$. 
\item \textit{Compositionality:}   following compositional distributional semantics models \cite{Trager2023LinearSO,PressZMSSL23,Stein24},  we focus on \textit{compositionality} through conjunctive (i.e., logical AND) concept composition \cite{naito2021revisiting}. The composition of \textit{atomic concepts} $C_1, \ldots, C_n$ yields a new \textit{composite concept} $C_1\times \ldots \times  C_n$, whose instances are formed by composing the instances of the constituent atomic concepts (e.g., $\langle$\textit{Prescot}, \textit{west\_of}$\land$\textit{far\_from}, \textit{Todmorden}$\rangle$).
\item \textit{Grounding}  is the awareness of  concepts, expressed using natural language constructs, with the physical entities in the world  they are assumed to apply to  \cite{Fodor1975-FODTLO, li2023can,yamada2024evaluating}. As done in previous work \cite{patel2022mapping}, we only use the \textit{text} modality  to represent concepts in two different modes. In our work, we specifically use numerical fact–based geometric measurements and associated question-based formulations.
\end{itemize}

\subsection{Concept probing methodology overview}
To  probe concept $C$ through the properties of \textit{abstraction}, \textit{compositionality}, and \textit{grounding}, we build a concept-centric question-answering benchmark  where questions $q \in \mathcal{Q}$ are generated from triplets $\langle x, r, y\rangle$ representing  instances of either atomic or composite concepts (\S~\ref{sec:bench}). 
Using this benchmark, we assess each concept property through specific tests (\S~\ref{prob}) by using two proxies:\\

\vspace{-0.4cm}
\noindent \textit{- Task performance:} to evaluate whether LLMs succeed on a test, we compute the \textit{accuracy} and \textit{consistency} metrics on the question-answering task. Let $\mathcal{Q}_t$ denote the question set. We define per-question \textit{accuracy} as $\mathrm{Acc}(q)=\mathbb{I}[\hat{y}(q)=y(q)]$, where $\hat{y}(q)$ is the LLM' prediction and $y(q)$ is the ground-truth answer. 
Accuracy is computed as the average over all questions $\mathcal{Q}_t$ as $\mathrm{Acc} = \frac{1}{|\mathcal{Q}_t|} \sum_{q \in \mathcal{Q}_t} \mathrm{Acc}(q)$. We access \textit{consistency}, regarding a concept property test, as a measure of whether a model correctly answers the paired questions $(q, \bar{q})$ that embed concept instances and their \textit{negation} through opposite relational facts (e.g., if $\langle x$, \textit{east\_of}, $y\rangle$ is true, then $\langle x$, \textit{west\_of}, $y\rangle$ should be false). Consistency is computed as $\mathrm{Consist.} = \frac{1}{|\mathcal{Q}_t|} \sum_{q \in \mathcal{Q}} \mathrm{Acc}(q) \cdot \mathrm{Acc}(\bar{q})$,  
where $\mathrm{Acc}(\bar{q})$ is the per-question accuracy of the negated question.\\
\noindent \textit{- Probing performance:}
We complement task performance evaluation with probing task evaluation \cite{belinkov-2022-probing}. The probing tasks rely on linear classifiers to test whether the internal  concept  instance representations $\mathbf{L}(q),  \,q \in \mathcal{Q}$ are aligned with the requirements of the property being tested.  

\section{The Concept-Centric Probing Benchmark}\label{sec:bench}

\subsection{Benchmarking geographic concepts}\label{subsec:bench_constr}
Without loss of generality, we investigate in this work three key  spatial concepts—
\emph{Direction}, \emph{Topology}, and \emph{Distance}. 
Let $\mathcal{W}$ be a set of geographic entities. We note $\mathcal{M}$  the set of wards that belong to UK metropolitan districts,\footnote{\label{fn:uk_metro_map}\href{https://en.wikipedia.org/wiki/Metropolitan_borough\#/media/File:English_metropolitan_boroughs_map_2021.svg}{Metropolitan boroughs on Wikipedia}: \url{en.wikipedia.org/wiki/Metropolitan_borough}.} $\mathcal{M} \in \mathcal{P(\mathcal{W})}$ where $\mathcal{P}$ denotes the power set. The distance, direction, and topology concepts can be expressed using a wide range of instances. In our work, we consider the following concept instances:
\begin{equation}
 \begin{aligned}
C_{dir} &= \{\text{north}, \text{south}, \text{east}, \text{west}\},\\
C_{dis} &= \{\text{close}, \text{far}\},\hspace{1em}
C_{top} = \{\text{within}, \text{borders}\}.
\end{aligned}   
\end{equation}
Each concept $C$ is represented using a set of instances $c$ as relational triplets 
$R_C = \{(x,r_c,y)/ (x,y) \in \mathcal{W} \times \mathcal{W}, c\in \mathcal{I}(C)  \}$. For the sake of simplicity, we consider $R_C=\cup_{c\in \mathcal{I}(C) } R_{c}$ (e.g., $R_{r_{dis}}= R_{r_{close}} \cup R_{r_{far}})$
To allow the evaluation of consistency, we also consider negated concept instance representations  $\{(x,\lnot{r_c},y)/ (x,y)  \in \mathcal{W} \times \mathcal{W}, c\in \mathcal{I}(C)  \}$. Table~\ref{tab:concept_relation_instance} presents the concepts, their instances, associated relations, and their negations.

\begin{table}[t]
\centering
\small
\resizebox{\columnwidth}{!}{
\begin{tabular}{l|c|c}
\hline
\textbf{Concept $C$} 
& \textbf{Instance ($c$) / Relation ($r_c$)} 
& \textbf{Negation ($\lnot r_c$)} \\
\hline
\multirow{2}{*}{Direction}
  & north/north\_of \;|\; south/south\_of
  & south\_of \;|\; north\_of \\
  & east/east\_of \;|\; west/west\_of
  & west\_of \;|\; east\_of \\
\hline
Distance
  & far/far\_from \;|\; close/close\_to
  & close\_to \;|\; far\_from \\
\hline
Topology
  & within/is\_within \;|\; borders/is\_bordering
  & is\_not\_within \;|\; is\_not\_bordering \\
\hline
\end{tabular}}
\caption{Concepts with paired instance--relation entries and their corresponding negated relations.}
\label{tab:concept_relation_instance}
\vspace{-1cm}
\end{table}

\subsection{GeoQA dataset Generation}\label{subsec:qa_gener_}

\paragraph{\textbf{Concept instance generation.}}  We focus on the UK metropolitan district wards\textsuperscript{\ref{fn:uk_metro_map}} as location entities, which form three discontinuous regions (upper, middle, and lower). Concept instances in the form of triplets $\langle x, r, y\rangle$ are generated from the middle continuous region $\mathcal{W}$, while the upper region is reserved for out-of-distribution (OOD) generalization. Specifically, using geometric measurements from GraphDB\footnote{https://graphdb.ontotext.com/} and YAGO2GEO\footnote{https://yago2geo.di.uoa.gr/}, we compute bearings, pairwise distances, and topological relations between wards associated with \textit{direction}, \textit{distance}, and \textit{topology} concepts, respectively. For any $x,y \in \mathcal{W}$, we denote their distance and bearing by $d(x,y)$ and $\theta(x,y)$, respectively, and map these measurements to relation concept instances $\langle x, r, y\rangle$ via the functions $\phi_{d,\mathrm{th}}$ and $\phi_{\theta}$, defined as follows: 
\vspace{-0.6cm}
\begin{figure}[h]
\centering
\scriptsize
\begin{minipage}{0.48\linewidth}
\[
\phi_{\theta}(x,y) =
\begin{cases}
\text{E}, & \theta \in [45^\circ, 135^\circ[,  \\ 
\text{S}, & \theta \in [135^\circ, 225^\circ[,  \\
\text{W}, & \theta \in [225^\circ, 315^\circ[,  \\
\text{N}, & \text{otherwise}. 
\end{cases}
\]
\end{minipage}
\hfill
\begin{minipage}{0.48\linewidth}
\[
\phi_{d,\mathrm{th}}(x,y) =
\begin{cases}
\text{close}, & d(x,y) \leq \mathrm{th},\\[2pt]
\text{far},   & d(x,y) > \mathrm{th}.
\end{cases}
\]
\end{minipage}
\vspace{-0.3cm}
\caption{Direction function $\phi_\theta$ mapping angles to cardinal directions. Distance threshold function $\phi_{d,\mathrm{th}}$ mapping distances to `close' or `far'.}
\label{fig:phi_functions}
\end{figure}

For example, $\phi_{\theta}(x,y)=\text{E}$ yields the triplet $(x,\text{east\_of},y)$, while
$\phi_{d,\mathrm{th}}(x,y)=\text{close}$ yields $(x,\text{close\_to},y)$; analogous triplets are generated for all remaining atomic distance and direction relations.  We select the distance threshold based on the pairwise distance distribution of geometric wards, setting $th$ as the mean of the distribution, resulting in $th = 47.76$ km. Since our goal is to investigate the internal LLM perception of distance, this threshold is not provided in the question context; its impact is further analyzed in Section~\ref{subsec:thresh}.
For topological relations, following YAGO2geo, the relations \textit{is\_within} and \textit{is\_bordering} between spatial entities are derived by applying GeoSPARQL/OGC topological predicates over their geometries. \\

The generation of all such atomic spatial triplets is described in Part~1 of Algorithm~\ref{alg:triplet_gen}(lines 4-16).
We further construct higher-level concept instances through \emph{conjunctive composition}, by systematically combining atomic relations that share the same subject and object. Given two atomic triplets $(x,r_1,y)$ and $(x,r_2,y)$, where $r_1$ and $r_2$ belong to different relational families (e.g., distance and direction), we generate a composite triplet $(x,r_1\land r_2,y)$ and proceed analogously for three relations. This composition process exhaustively covers all valid pairwise combinations of atomic relations, and is further extended to higher-order compositions (i.e., triplets of relations) whenever applicable. \textbf{By explicitly generating atomic, pairwise, and higher-order composed relations, the benchmark enables controlled evaluations of abstraction and compositional generalization}.
The compositional generation procedure is detailed in Part~2 of Algorithm~\ref{alg:triplet_gen}(lines 17-22).

\paragraph{\textbf{Question generation.}}
We generate Binary-QA (Yes/No) and MCQ (3-options) from each concept instance represented as triplets generated by
Algorithm~\ref{alg:triplet_gen}. Given the resulting set of  positive triplets
$T^{+}$, we construct negative triplets $T^{-}$ by negating each relation $r$ (Table~\ref{tab:concept_relation_instance}), yielding the complete set $\mathcal{T} = T^{+} \cup T^{-}$. We then apply a predefined set of templates to each triplet $t \in \mathcal{T}$, illustrated in Table~\ref{tab:question_examples}. The prompt templates used for binary and MCQ tasks are defined are presented in Figure~\ref{fig:prompt_templates}.

\begin{figure}[h]
    \centering
    \begin{tcolorbox}[
        colback=cyan!10!white,
        colframe=cyan!70!black,
        boxsep=0pt,
        left=2pt, right=2pt, top=2pt, bottom=2pt,
        sharp corners
    ]
    \noindent\textbf{Binary-QA Task:} Answer the following geography question on UK metropolitan district wards. Respond only with \texttt{yes} or \texttt{no}.\\
    \noindent\textbf{MCQ Task:} Answer the following geography question on UK metropolitan district wards. Respond only with the correct option A, B, or C.
    \end{tcolorbox}
    \vspace{-0.4cm}
    \caption{Prompt templates for Binary-QA and MCQ tasks.}
    \label{fig:prompt_templates}
    \vspace{-0.4cm}
\end{figure}

\paragraph{\textbf{Ground-truth answer generation}.}
Ground-truth answers for both, Binary-QA and MCQ tasks, are generated using Algorithm~\ref{alg:gt_answers}. 
MCQ tasks additionally require the generation of incorrect options, which are produced using a distractor generator denoted by $D_n$:
\[
D_n(R,y)=\{d_1,\dots,d_n \mid d_i\sim\{x\in\mathcal{W}\mid (x,y)\notin R\}\}.
\]
In our case $n=2$, the procedure is to choose the target $w^\ast\in\mathcal{W}$ as the correct answer, sample two distinct distractors $w_1,w_2\in\mathcal{W}\setminus\{w^\ast\}$ satisfying $(w_i,y)\notin R$, form the option set $O=\{w^\ast,w_1,w_2\}$, and place $w^\ast$ at a uniformly random position $i^\ast\in\{1,2,3\}$.

\begin{algorithm}[t]
\caption{Triplet Generation}
\label{alg:unified_triplet_generation}
\begin{algorithmic}[1]
\State \textbf{Inputs:} wards $\mathcal{W}$, metropolitan districts $\mathcal{M}$
\State \textbf{Output:} $T^+$ (atomic and compositional positive triplets)
\State Initialize $R_{r_{dis}},R_{r_{dir}},R_{r_{top}},R_{{is\_within}},T^+\gets\varnothing$

\ForAll{$x,y\in\mathcal{W}$}  \Comment{\textcolor{blue}{Part 1: Atomic triplet}}
    \State  $d\gets d(x,y)$, $\theta\gets\theta(x,y)$ \Comment{\textcolor{DarkGreen}{compute distance and angle}}
    \State $r_{dis}\gets\phi_d(x,y)$,  $r_{dir}\gets\phi_\theta(x,y)$  \Comment{\textcolor{DarkGreen}{dis. and dir. relation}}
    \State $R_{r_{dis}}\gets R_{r_{dis}}\cup\{(x,y)\}$;\quad
           $R_{r_{dir}}\gets R_{r_{dir}}\cup\{(x,y)\}$
    \If{border$(x,y)$} \Comment{\textcolor{DarkGreen}{Using YAGO2GEO}}
         \State $r_{top} \gets is\_bordering$
        \State $R_{r_{top}}\gets R_{r_{top}}\cup\{(x,y)\}$
    \EndIf
    \State $r_{top} \gets is\_within$
    \State Find $z \in \mathcal{M}$, such that $(x, r_{top}, z)$ \Comment{\textcolor{DarkGreen}{Using YAGO2GEO}}
    \State $R_{r_{top}} \leftarrow R_{r_{top}} \cup \{(x,z)\}$
    \State $T^+ \gets T^+ \cup  \{(x,r, y) \quad \forall r \in \{r_{dir},r_{dis}, r_{top}\} \}$
\EndFor

\ForAll{$(x,y) \in T^+$}\Comment{\textcolor{blue}{Part 2: Compositional triplet }}
    \State $S \gets \{ r \mid (x,r,y)\in T^+\}$ 
    \ForAll{subsets $U\subseteq S$ with $|U|\ge 2$} 
        \State $r_U \gets \bigwedge_{r\in U} r$ \Comment{logical AND (e.g.\ $r_1\land r_2$)}
        \State $T^+ \gets T^+ \cup \{(x,r_U,y)\}$
    \EndFor
\EndFor

\State \Return $T^+$
\end{algorithmic}
\label{alg:triplet_gen}
\end{algorithm}

\begin{algorithm}[t]
\caption{Ground Truth Generation of Answers (A)}
\label{alg:ground_truth}
\begin{algorithmic}[1]
\State \textbf{Inputs:}  
$k \in \{\text{Binary-QA}, \text{MCQ}\}$
,$T^{+}$, $\mathcal{W}$, $D_n$
\State \textbf{Outputs:} Answer set $A$
\For{all $(x,R,y) \in T^{+}_{\mathcal{R}^{+}} $} \Comment{\textcolor{blue}{Binary task answers}}
        \If{$k = \text{Binary-QA}$}
                \State $A(x, R , y) \gets \{\text{Yes}\}$;\quad $A(x,\lnot{R}, y) \gets \{\text{No}\}$
        \ElsIf{$k = \text{MCQ}$} \Comment{\textcolor{blue}{MCQ task answers}}
                 \State $A(x, R , y) \gets \{x,d_1,d_2\}$ (i.e. ${x} \cup D_2(R, y)$)
                 
                 \State $\overline{x} \sim \{ z \in \mathcal{W} \mid (z,y) \in  \lnot{R} \}$ \Comment{\textcolor{DarkGreen}{Sampling ($\sim$)}}
                 
                 \State $A(x, \lnot{R} , y) \gets \{\overline{x},\overline{d_1},\overline{d_2}\}$ (i.e. ${x} \cup D_2(\lnot{R}, y)$)     

                \State $A(y) \gets \overline{x}$
        \EndIf
\EndFor
\State \Return $A$
\end{algorithmic}
\label{alg:gt_answers}
\end{algorithm}

\begin{table}[t]
\centering

\resizebox{\columnwidth}{!}{
\begin{tabular}{l|l|l}
\hline
\textbf{Affirmation} 
& \textbf{Negation} 
& \textbf{Composition Example} \\ \hline

Is $X$ far from $Y$?            
& Is $X$ close to $Y$? 
& Is $X$ far from $Y$ \textbf{and} close to $Y$? \\

Is $X$ close to $Y$?            
& Is $X$ far from $Y$? 
& Is $X$ close to $Y$ \textbf{and} far from $Y$? \\

Is $X$ west of $Y$?        
& Is $X$ east of $Y$? 
& Is $X$ west of $Y$ \textbf{and} bordering $Y$? \\

Does $X$ border $Y$?            
& Is $X$ not bordering $Y$? 
& Does $X$ border $Y$ \textbf{and} north of $Y$? \\

Is $X$ within $Z$?              
& Is $X$ not within $Z$? 
& Is $X$ within $Z$ \textbf{and} close to $Y$? \\ \hline
\end{tabular}}
\caption{Examples of binary atomic and compositional questions, for MCQ task, we use slightly different template (e.g., \textit{Which ward is close to $Y$?} followed by options).}
\label{tab:question_examples}
\vspace{-.8cm}
\end{table}

\paragraph{\textbf{QA-Dataset statistics.}} Let $|.|$ denotes the cardinality measure, we have $|\mathcal{W}| = 506$ wards and $|\mathcal{M}| = 25$ metropolitan district. Statistics of our generated Geo-QA dataset, \(\mathcal{D}_{QA}\), are presented in Table~\ref{tab:data_stats}.

\begin{table}[h]
\centering
\small
\resizebox{0.9\columnwidth}{!}{
\begin{tabular}{llr}
\toprule
Task(s) & Combinations & Total \\
\midrule
\multirow{4}{*}{Binary-QA} 
 & 1-Concept: dir, dist, top  & 769{,}260 \\
 & 2-Concept: dir\_top, dist\_top , dist\_dir & 771{,}930 \\
 & 3-Concept: dist\_dir\_top  & 258{,}200 \\ \cline{2-3}
 & \textbf{Overall Total} & 1{,}799{,}390 \\
\midrule
\multirow{4}{*}{MCQ} 
 & 1-Concept: dir, dist, top & 4{,}096 \\
 & 2-Concept: dir\_top, dist\_top, dist\_dir & 70{,}100 \\
 & 3-Concept: dist\_dir\_top & 41{,}098 \\ \cline{2-3}
 & \textbf{Overall Total} & 115{,}294 \\
\bottomrule
\end{tabular}}
\caption{Statistics of the Geo-QA dataset ($\mathcal{D}_{QA}$) per task for Binary-QA and MCQ.}
\label{tab:data_stats}
\vspace{-1cm}
\end{table}

\section{Probing LLMs on geo-spatial QA}\label{subsec:prelime_eval} Our key objective here is to investigate whether geo-QA task performance is a reliable and sufficient proxy to probe LLMs on their understanding of concepts.
\subsection{Preliminary evaluation}\label{subsec:prelime_eval}

Before testing LLM's understanding of concepts on core properties, we first aim to evaluate their performance on the QA dataset of our benchmark, \(\mathcal{D}_{QA}\) (§ Table~\ref{tab:data_stats}).  We evaluate popular text-only LLMs (Llama-8B, Mistral-7B/8B, Qwen-0.6B, 1.7B, 4B, 8B)  and report the main results in Table~\ref{tab:binary_mcq_shared}. We can see that across all settings, MCQ accuracy consistently exceeds Binary-QA accuracy, confirming that structured answer choices substantially ease the task. Larger instruction-tuned models achieve the highest accuracies (e.g., Llama-3.1-8B at \(71.7\%\) MCQ and Qwen3-4B at \(56.14\%\) Binary-QA). However, consistency remains low overall, especially for Binary-QA (often \(<26\%\)), indicating unstable factual recall despite moderate accuracy. 
While explicit thinking is costly, we can observe that it is largely ineffective: it yields marginal gains at best and often degrades performance (e.g., LLaMA-3.1-8B MCQ accuracy drops from \(71.7\%\) to \(50.6\%\), with consistency decreasing from \(48.7\%\) to \(16.6\%\)). Smaller models sometimes reach comparable or slightly improved accuracy under explicit thinking but still suffer noticeable consistency declines, suggesting correct answers without robust consistency. Consequently, we focus henceforth on models without explicit thinking, relying instead on their inner latent reasoning for our different tests on LLMs' concept understanding.

\begin{figure}[ht]
    \centering
    \includegraphics[width=\columnwidth]{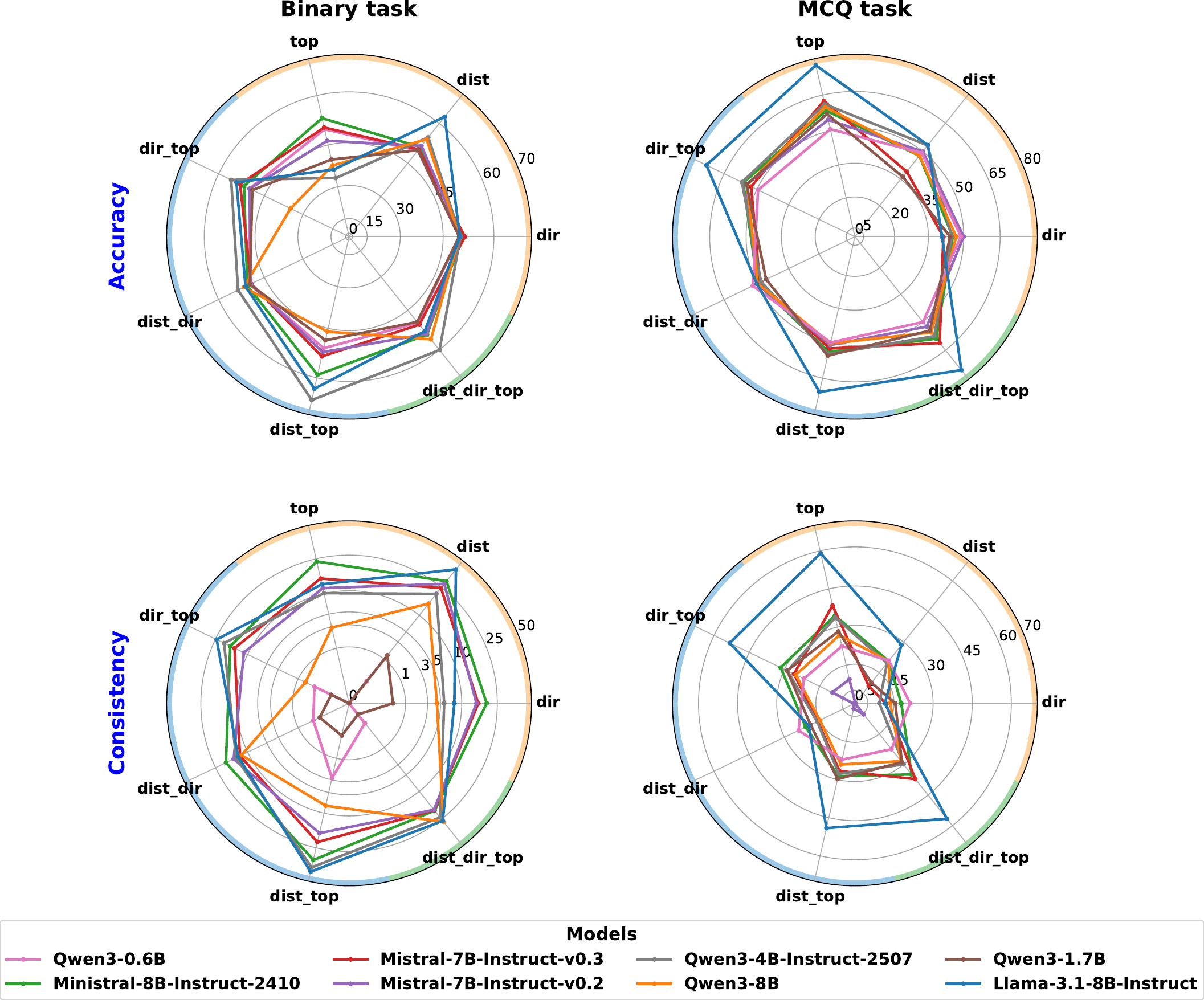}
    \caption{Performance trends per concept type and composition from one to three concepts across Binary-QA and MCQ tasks.}
    \label{fig:combined_trends}
    \vspace{-0.8cm}
\end{figure}

Focusing on concept-level analysis, Figure~\ref{fig:combined_trends} highlights performance across concept dimensions. In particular, we can see that the LLaMA3-8B model exhibits relatively higher accuracy across most settings compared to other models, reaching above $70\%$  on the $dist\_dir\_top$ composition, with broader coverage across the seven dimensions representing the full concept space. However, consistency remains comparatively lower, reaching around $60\%$ at best on $top$ concept (MCQ task). Overall, models tend to perform better under standard prompting than with explicit reasoning, with compositional settings not consistently outperforming atomic ones in the MCQ task.

\definecolor{mycyan}{RGB}{0,160,160}

\begin{table}[t]
\centering
\resizebox{\columnwidth}{!}{
\begin{tabular}{l
c >{\columncolor{mycyan!30}}c
c >{\columncolor{mycyan!30}}c|
c >{\columncolor{mycyan!30}}c
c >{\columncolor{mycyan!30}}c}
\hline
\textbf{Model/Task} & \multicolumn{4}{c|}{\textbf{Binary\_QA}} & \multicolumn{4}{c}{\textbf{MCQ}} \\
\cline{2-5} \cline{6-9}
 & \multicolumn{2}{c}{Acc (\%)} & \multicolumn{2}{c|}{Consist. (\%)} & \multicolumn{2}{c}{Acc (\%)} & \multicolumn{2}{c}{Consist. (\%)} \\ \cline{2-8}
 & $\checkmark$ & $\times$ & $\checkmark$ & $\times$ & $\checkmark$ & $\times$ & $\checkmark$ & $\times$ \\
\hline

Llama-3.1-8B-Inst. 
& 61.58 & 54.95 & 40.04 & 25.68 & 50.60 & \textbf{71.70} & 16.61 & \textbf{48.72} \\

Mistral-2410-8B-Inst. 
& 61.26 & 53.39 & 38.22 & 22.94 & 49.10 & 56.16 & 23.73 & 30.96 \\

Mistral-7B-v0.3-Inst. 
& 51.64 & 51.65 & 22.55 & 18.09 & 51.87 & 55.31 & 21.63 & 31.64 \\

Mistral-7B-v0.2-Inst. 
& 55.42 & 51.28 & 27.24 & 17.10 & 58.54 & 52.73 & 22.40 & 32.32 \\

Qwen3-8B 
& 62.30 & 47.99 & 41.02 & 10.83 & 47.22 & 53.52 & 15.96 & 24.25 \\

Qwen3-4B-Inst. 
& 62.73 & 56.22 & 39.98 & 20.81 & 50.00 & 56.14 & 22.67 & 25.05 \\

Qwen3-1.7B 
& 56.05 & 49.09 & 30.74 & 0.30 & 46.17 & 54.29 & 19.47 & 28.54 \\

Qwen3-0.6B 
& 53.52 & 50.25 & 19.86 & 0.62 & 49.06 & 49.90 & 21.09 & 22.02 \\

\hline
\end{tabular}}
\caption{MCQ and Binary-QA performance with shared metrics. \checkmark indicates explicit thinking (reasoning-enabled for Qwen models and CoT for the others); \ding{55} indicates standard prompting.}
\vspace{-0.6cm}
\label{tab:binary_mcq_shared}
\end{table}

\begin{table*}[t]
\centering
\setlength{\tabcolsep}{4pt}
\renewcommand{\arraystretch}{0.92}
\footnotesize
\begin{tabular}{@{}lrrrrrrrrrrrr@{}}
\toprule
 & \multicolumn{6}{c}{\textbf{UK} ($T_0=47.76$\,km)} & \multicolumn{6}{c}{\textbf{US} ($T_0=1636.66$\,km)} \\
\cmidrule(lr){2-7}
\cmidrule(lr){8-13}
 & \multicolumn{3}{c}{\textbf{Binary}} & \multicolumn{3}{c}{\textbf{MCQ}} & \multicolumn{3}{c}{\textbf{Binary}} & \multicolumn{3}{c}{\textbf{MCQ}} \\
\cmidrule(lr){2-4}
\cmidrule(lr){5-7}
\cmidrule(lr){8-10}
\cmidrule(lr){11-13}
\textbf{Model} & $T^*$ & Acc\,(\%) & Cons\,(\%) & $T^*$ & Acc\,(\%) & Cons\,(\%) & $T^*$ & Acc\,(\%) & Cons\,(\%) & $T^*$ & Acc\,(\%) & Cons\,(\%) \\
\midrule
Llama-3.1-8B & 16.80 & 70.0 \phantom{0}(\textcolor{blue}{+7.6}) & 48.3 \phantom{0}(\textcolor{blue}{+7.6}) & 43.22 & 41.0 \phantom{0}(\textcolor{blue}{+2.7}) & 13.4 \phantom{0}(\textcolor{blue}{+3.0}) & 1824.39 & 48.8 \phantom{0}(\textcolor{red}{-2.9}) & 25.4 \phantom{0}(\textcolor{red}{-2.9}) & 1365.79 & 43.4 (\textcolor{blue}{+1.7}) & 0.0 (\textcolor{black}{+0.0}) \\
Ministral-8B & 20.88 & 60.3 \phantom{0}(\textcolor{blue}{+9.0}) & 38.8 \phantom{0}(\textcolor{blue}{+9.0}) & 43.99 & 39.8 \phantom{0}(\textcolor{blue}{+2.4}) & 16.0 \phantom{0}(\textcolor{blue}{+1.6}) & 1225.86 & 58.5 \phantom{0}(\textcolor{blue}{+8.4}) & 38.0 \phantom{0}(\textcolor{blue}{+8.4}) & 1570.75 & 31.6 (\textcolor{blue}{+0.5}) & 0.0 (\textcolor{black}{+0.0}) \\
Mistral-7B-v0.2 & 20.78 & 65.3 (\textcolor{blue}{+14.5}) & 41.5 (\textcolor{blue}{+14.5}) & 56.36 & 67.6 (\textcolor{blue}{+17.0}) & 22.2 (\textcolor{blue}{+11.1}) & 1034.62 & 64.2 (\textcolor{blue}{+13.4}) & 47.2 (\textcolor{blue}{+13.5}) & 1495.82 & 35.9 (\textcolor{red}{-2.8}) & 14.3 (\textcolor{black}{+0.0}) \\
Mistral-7B-v0.3 & 19.70 & 64.4 (\textcolor{blue}{+14.0}) & 38.3 (\textcolor{blue}{+14.0}) & 28.05 & 44.2 \phantom{0}(\textcolor{red}{-1.0}) & 16.8 \phantom{0}(\textcolor{red}{-2.0}) & 344.80 & 80.0 (\textcolor{blue}{+30.8}) & 63.6 (\textcolor{blue}{+30.8}) & 1486.11 & 32.5 (\textcolor{black}{+0.0}) & 14.3 (\textcolor{blue}{+4.8}) \\
Qwen3-0.6B & 47.37 & 50.0 \phantom{0}(\textcolor{black}{+0.0}) & 0.1 \phantom{0}(\textcolor{black}{+0.0}) & 45.09 & 35.9 \phantom{0}(\textcolor{blue}{+2.4}) & 11.9 \phantom{0}(\textcolor{blue}{+0.2}) & 1293.36 & 50.1 \phantom{0}(\textcolor{blue}{+0.1}) & 2.0 \phantom{0}(\textcolor{blue}{+0.1}) & 1508.52 & 32.9 (\textcolor{blue}{+0.7}) & 0.0 (\textcolor{black}{+0.0}) \\
Qwen3-1.7B & 7.54 & 50.1 (\textcolor{black}{+0.0}) & 1.3 (\textcolor{black}{+0.0}) & 44.43 & 35.5 \phantom{0}(\textcolor{blue}{+1.8}) & 11.9 \phantom{0}(\textcolor{blue}{+2.0}) & 1724.44 & 50.3 \phantom{0}(\textcolor{red}{-0.4}) & 10.7 \phantom{0}(\textcolor{red}{-0.4}) & 1604.81 & 24.7 (\textcolor{red}{-0.3}) & 0.0 (\textcolor{black}{+0.0}) \\
Qwen3-4B & 30.86 & 55.9 \phantom{0}(\textcolor{blue}{+0.9}) & 21.6 \phantom{0}(\textcolor{blue}{+0.9}) & 43.65 & 35.6 \phantom{0}(\textcolor{blue}{+2.7}) & 14.3 \phantom{0}(\textcolor{blue}{+1.2}) & 1055.08 & 60.0 \phantom{0}(\textcolor{blue}{+9.5}) & 32.1 \phantom{0}(\textcolor{blue}{+9.5}) & 1503.55 & 36.9 (\textcolor{blue}{+2.8}) & 15.6 (\textcolor{black}{+0.0}) \\
Qwen3-8B & 31.74 & 57.1 \phantom{0}(\textcolor{blue}{+2.9}) & 18.1 \phantom{0}(\textcolor{blue}{+2.9}) & 44.89 & 36.8 \phantom{0}(\textcolor{blue}{+1.9}) & 11.6 \phantom{0}(\textcolor{blue}{+0.6}) & 1582.63 & 52.8 \phantom{0}(\textcolor{blue}{+0.9}) & 28.0 \phantom{0}(\textcolor{blue}{+0.9}) & 1453.31 & 37.5 (\textcolor{blue}{+1.1}) & 0.0 (\textcolor{black}{+0.0}) \\
\bottomrule
\end{tabular}
\caption{Revealed threshold $T^*$ (km) and performance relative to the GT label $T_0$. Acc\,(\%) and Cons\,(\%) are evaluated at $T^*$; deltas in parentheses show gain over evaluation at $T_0$ (\textcolor{blue}{$+$}\,=\,improvement, \textcolor{red}{$-$}\,=\,degradation).}
\label{tab:threshold}
\vspace{-0,5 cm}
\end{table*}

\subsection{Distance threshold analysis} \label{subsec:thresh}
In this part, we investigate the internal definition of the distance concept per LLMs, regions and tasks. Binary and MCQ distance labels are derived from a fixed threshold~$T_0$ set to the pairwise distance median. A key question is whether $T_0$ reflects the models' \emph{implicit} notion of
closeness or imposes an external boundary misaligned with LLM perception. Moreover, to evaluate the effect of \textbf{regional scale}, we regenerate the data similarly as UK for the US region. For each region, and task, we estimate each model's \textbf{revealed threshold}~$T^*$ as the intersection of the kernel density estimation (KDE) curves fitted  to the close- and far-labelled distance distributions extracted from existing predictions. For binary questions, question polarity combined with Yes/No prediction, determines question label, for MCQ, the distance to the \emph{chosen} ward is assigned with question relation (close/far). Results are summarized in Table~\ref{tab:threshold}. We highlight three main findings:
\noindent\textbf{(i) Closeness perception is systematically stricter than the dataset label.}
In the binary task, most models yield $T^*_{\mathrm{UK}} \ll T_0$, treating a large fraction of \emph{close}-labeled pairs as \emph{far}. This compression is consistent across families, suggesting a corpus-level prior on proximity rather than a model-specific artifact.
\noindent\textbf{(ii)~Task format re-calibrates the implicit threshold.} MCQ reverses the binary pattern: all models cluster near
$T^*/T_0 {\approx} 0.92$--$0.94$ (UK), closely tracking~$T_0$.
The three named candidate wards act as implicit distance anchors,
pulling the model's boundary toward the dataset scale, regardless of its
underlying spatial prior.
This re-calibration is format-induced rather than evidence of stronger
spatial understanding, and cautions against interpreting MCQ accuracy
as a reliable proxy for absolute closeness concept strength.\noindent\textbf{(iii)~Distance concept understanding degrades at continental scale.} Smaller Qwen models (0.6B, 1.7B) show near-chance US binary accuracy ($\approx 50\%$) with negligible~$\Delta$, indicating no close/far discrimination at the 1{,}637~km scale, whereas larger models maintain meaningful gains (Qwen3-4B: Acc\,+6.1, Cons\,+5.3; Qwen3-8B: Acc\,+5.7)---suggesting continental-scale distance concepts are capacity-dependent. The same small models yield non-trivial improvements in UK binary (Qwen3-4B: Acc\,+3.2, Cons\,+4.8), confirming the collapse is scale-induced rather than a general model limitation. This degradation manifests differently in MCQ: nearly all models yield \emph{negative} accuracy deltas at $T^*$ across both regions (e.g.\ Qwen3-1.7B US: $-2.1$, Qwen3-4B UK: $-0.5$), meaning the fixed label $T_0$ already outperforms the model's own revealed boundary in the anchored-option setting---further evidence that MCQ performance reflects format calibration rather than intrinsic spatial understanding.

\textbf{To sum up.} QA-based evaluation has inherent limitations as a proxy for concept understanding: surface-level accuracy does not reliably reflect robust understanding across tasks and regions, particularly for challenging concepts such as distance.
Consequently, a more targeted probing approach is required to analyze the core properties of concepts in depth, which is the central focus of the following Section~\ref{prob}.
\section{Concept Probing  }\label{prob}

 To evaluate concept \textit{abstraction}, \textit{compositionality}, and \textit{grounding}, we construct a dedicated \textbf{probing dataset} \(\mathcal{D}_{PB}\) by subsampling 1{,}000 binary and 1{,}000 ternary representative compositional questions from our Geo-QA corpus \(\mathcal{D}_{QA}\) and including all corresponding atomic decompositions and their negations, resulting in $2\times7{,}000=14{,}000$ QA instances. Particularly for the MCQ task, we ensure that each compositional question and its atomics share the same options, this is mandatory for consistent decomposition. Statistics of \(\mathcal{D}_{PB}\) are presented in Table~\ref{tab:combination_counts_transposed}.

\begin{table}[t]
\centering
\resizebox{\columnwidth}{!}{
\begin{tabular}{l c c c c c c c}
\hline
\multicolumn{7}{c}{Probing Data $(\mathcal{D}_{PB})$: Concept types }\\
\hline
\textbf{dir} & \textbf{dist} & \textbf{dist\_dir\_top} & \textbf{top} & \textbf{dir\_top} & \textbf{dist\_top} & \textbf{dist\_dir} \\
\hline
1541 & 1523 & 1000 & 844 & 374 & 354 & 272 \\
\bottomrule
\end{tabular}}
\caption{Positive counts for each combination in the probing data with uniform sampling per number of concepts.}
\label{tab:combination_counts_transposed}
\vspace{-0.8cm}
\end{table}

\vspace{-0.2cm}
\subsection{Testing concept abstraction }
\label{subsec:abstraction}
\tcbset{
    myheader/.style={
        colback=orange!10,    
        colframe=orange!70,   
        coltitle=black,     
        fonttitle=\bfseries,
        enhanced,
        boxrule=0.8mm,
        arc=4mm,
        left=2mm,
        right=2mm,
        top=1mm,
        bottom=1mm
    }
}

To test LLMs on the property of \textit{abstraction}, we answer two core questions tightly related to semantic type representativeness and generalizability: (i) Are concept instances explicitly encoded in the hidden layer and reliably classified as concept types? (ii) Do learned concept representations generalize to unseen instances or novel combined instances? \\
\noindent\textbf{Experimental design.} To investigate whether frozen LLMs encode geographical concepts at different depths, we employ a layer-wise linear probing approach. For each layer $l$ of a pre-trained model, we extract the average token embedding for a question $q$, producing a layer-specific representation $h = f_l(q)$. A probe classifier trained on $h$ predicts one of seven conceptual classes $C$ (e.g., directional, topological, distance, or their compositions). Specifically, for each $q$, the probe outputs $\hat{y} = \mathrm{softmax}(W_l h + b_l)$
where $c_q \in C$ is the predicted label and $W_l, b_l$ are layer-specific parameters.

To evaluate the generalizability of LLM representations, we construct several data splits: a standard random train/test split to probe concept encoding (question (i)), and two OOD settings,  to probe generalization and \emph{systematicity} \cite{Fodor1975-FODTLO} (question (ii)): the region-based OOD uses QA pairs from the \textit{upper-region}, while the token-based OOD withholds specific concept combinations. 
All splits are summarized in Table~\ref{tab:ood_splits}.


\begin{table}[h]
\centering
\small
\resizebox{\columnwidth}{!}{
\begin{tabular}{lcc}
\toprule
\textbf{Split Type} & \textbf{Train} & \textbf{Test} \\
\midrule

\multicolumn{3}{c}{\textit{Token-Level Split (Atomic-Based)}} \\
\midrule
Direction & \{N, S\} & \{E, W\} \\
Distance  & \{C\}    & \{F\} \\
Topology  & \{Wi\}   & \{B\} \\

\midrule
\multicolumn{3}{c}{\textit{Combination-Based}} \\
\midrule
Atomics & 50\% per token & 50\% per token \\
Dis$\times$Dir & \{C, F\}$\times$\{N, S\} & \{C, F\}$\times$\{E, W\} \\
Dis$\times$Top & \{C$\times$Wi, F$\times$B\} & \{C$\times$B, F$\times$Wi\} \\
Dir$\times$Top & \{N$\times$Wi, S$\times$B\} & \{E$\times$Wi, W$\times$B\} \\
Dis$\times$Dir$\times$Top & \{C, F\}$\times$\{N, S\}$\times$\{Wi, B\} & \{C, F\}$\times$\{E, W\}$\times$\{Wi, B\} \\

\midrule
\multicolumn{3}{c}{\textit{Geo-Level Split\textsuperscript{\ref{fn:uk_metro_map}}}} \\
\midrule
Geo-OOD & \{mid-region\} & \{upper-region\} \\

\bottomrule
\end{tabular}
}
\caption{Atomic-based token-level and combination-based compositional OOD splits, followed by geographic OOD evaluation.
\textbf{Token abbreviations:} N=North, S=South, E=East, W=West, C=Close, F=Far, Wi=Within, B=Borders.}
\label{tab:ood_splits}
\vspace{-.8cm}
\end{table}

\begin{figure}[ht]
    \centering
    \includegraphics[width=0.9\columnwidth]{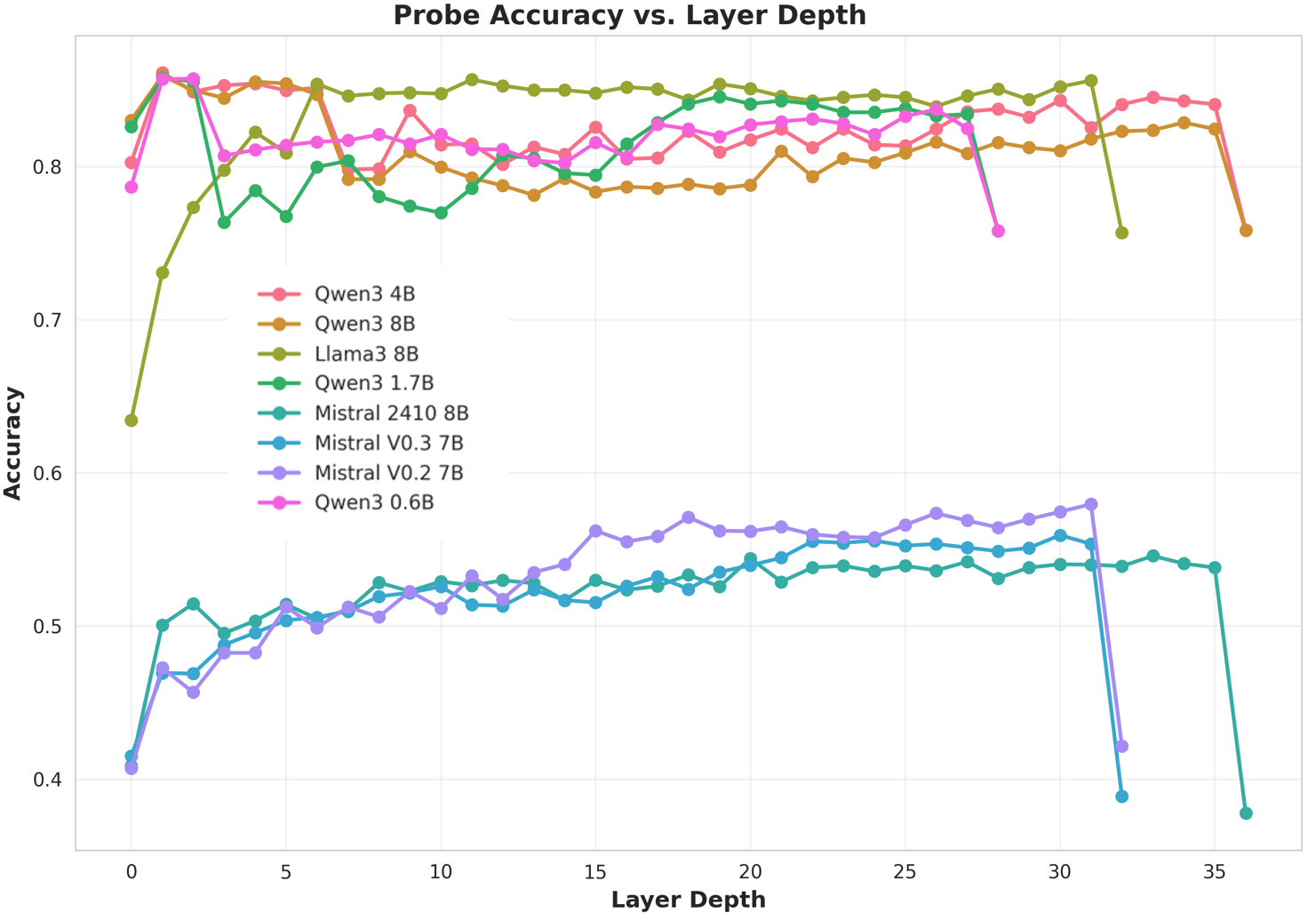}
    \caption{Binary Task: Concepts classification accuracy per LLM layer (depth) for Geo-Level split. The other splits shows similar trends for both tasks.} 
    \label{fig:cls_concepts_prob}
    \vspace{-0.5 cm}
\end{figure}

\noindent\textbf{Results.}  Table~\ref{tab:final_probe_overall} and Figure~\ref{fig:cls_concepts_prob} show that most evaluated LLMs achieve near-perfect performance on the standard random split (\(99.95\!-\!99.98\%\) accuracy) and generalize well to OOD settings, reaching \(75\!-\!83\%\) on the Geo-Level split, \(80\!-\!83\%\) on the single Token-level split, and over \(99\%\) on the compositional Token-level split. In contrast, the Mistral-family models exhibit substantially lower accuracy across all splits --random split $52.8\!-\!58.0\%$, Geo-Level split \(37.8\!-\!42.2\%\), single Token-level split \(22.5\!-\!29.4\%\), and compositional Token-level split \(48.5\!-\!55.4\%\)-- indicating significantly weaker concept encoding and generalization under comparable experimental settings.

\noindent\textbf{To sum up}: Most models clearly recognize concepts via their latent representations, which is a consistent finding with previous work that has shown the ability of LLMs to represent the space through the geographic entity location task \cite{gurnee2024language}. However, the Mistral family consistently underperforms across all splits and layers, suggesting limited concept abstraction capacity in its architecture compared to the other tested LLMs.

\begin{table}[t]
\centering
\small
\resizebox{\columnwidth}{!}{
\begin{tabular}{l c c cc c}
\toprule
 &  &  & \multicolumn{2}{c}{\textbf{Token-Level Test}} & \textbf{Geo-Level Test} \\
\cmidrule(lr){4-5}
\textbf{Model} & \textbf{L} & \textbf{Random} & \textbf{Single-Based} & \textbf{Combination-Based} & \textbf{Geo} \\
\midrule
qwen3\_4b         & 36 & \cellcolor{green!30}99.98 & \cellcolor{green!30}81.65 & \cellcolor{green!30}99.45 & \cellcolor{green!30}75.86 \\
qwen3\_8b         & 36 & \cellcolor{green!30}99.98 & \cellcolor{green!30}80.76 & \cellcolor{green!30}99.61 & \cellcolor{green!30}75.83 \\
llama3\_8b        & 32 & \cellcolor{green!30}99.95 & \cellcolor{green!30}82.85 & \cellcolor{green!30}99.90 & \cellcolor{green!30}75.69 \\
qwen3\_1.7b       & 28 & \cellcolor{green!30}99.98 & \cellcolor{green!30}80.43 & \cellcolor{green!30}99.10 & \cellcolor{green!30}75.81 \\
mistral\_2410\_8b & 36 & \cellcolor{red!30}53.05  & \cellcolor{red!30}22.47 & \cellcolor{red!30}48.47 & \cellcolor{red!30}37.76 \\
mistral\_v0.3\_7b & 32 & \cellcolor{red!30}52.83  & \cellcolor{red!30}24.17 & \cellcolor{red!30}49.15 & \cellcolor{red!30}38.88 \\
mistral\_v0.2\_7b & 32 & \cellcolor{red!30}58.02  & \cellcolor{red!30}29.40 & \cellcolor{red!30}55.37 & \cellcolor{red!30}42.17 \\
qwen3\_0.6b       & 28 & \cellcolor{green!30}99.95 & \cellcolor{green!30}80.87 & \cellcolor{green!30}99.33 & \cellcolor{green!30}75.81 \\
\bottomrule
\end{tabular}}
\caption{Final-layer probe results across evaluation settings.
\textbf{Single-Based} and \textbf{Combination-Based} correspond to atomic and compositional token-level OOD tests, respectively.
\textbf{Geo-Level Test} evaluates geographic generalization.}
\label{tab:final_probe_overall}
\vspace{-.8cm}
\end{table}

\begin{figure*}
    \centering
    \small
    \begin{subfigure}[b]{\columnwidth}
        \centering
        \includegraphics[width=\columnwidth,trim=0cm 0.7cm 0cm 0cm, clip]{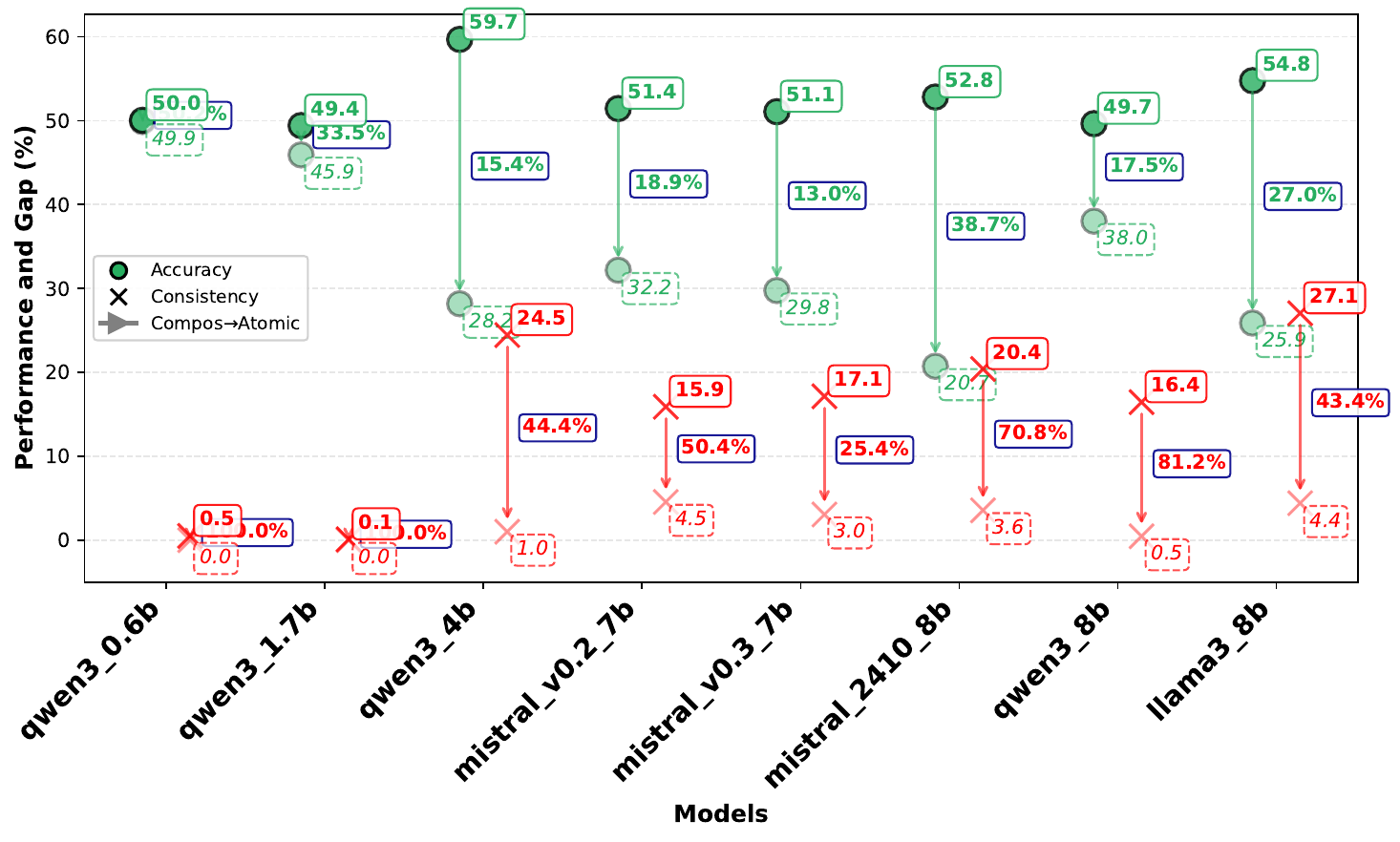}
        \caption{Binary-QA Task.}
        \label{fig:binary_compos}
    \end{subfigure}
    \hfill
    \begin{subfigure}[b]{\columnwidth}
        \centering
        \includegraphics[width=\columnwidth,trim=0cm 0.7cm 0cm 0cm, clip]{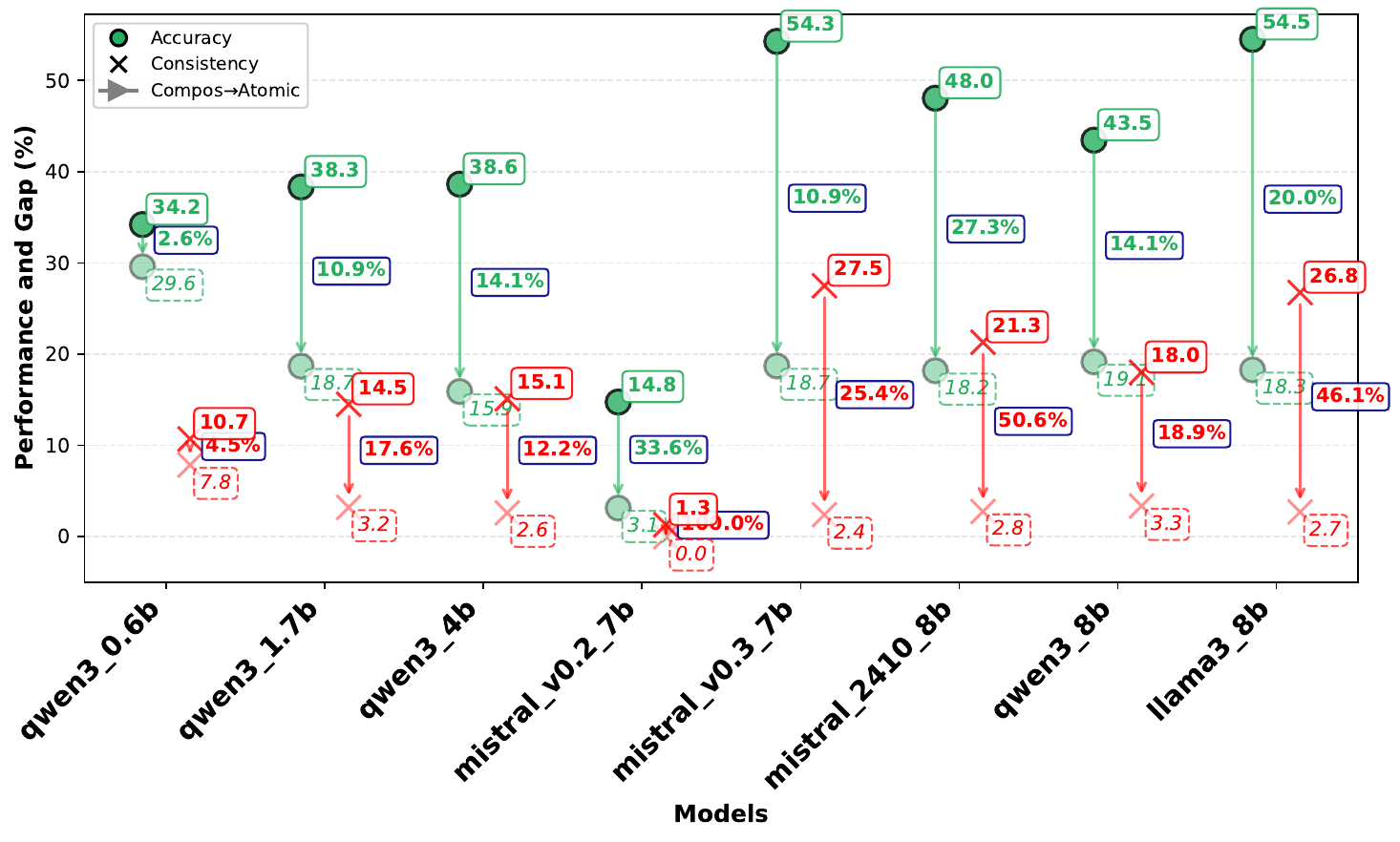} 
        \caption{MCQ Task.}
        \label{fig:compositional_gap}
    \end{subfigure}
    \vspace{-0.3cm}
    \caption{Compositional gap analysis (in blue rectangle), highlighting differences between composite and atomic question performance w.r.t accuracy  of compositional questions and their atomics. The arrow points from compositional to atomics, illustrating the gap.}
    \label{fig:merged_compos}
\end{figure*}
\subsection{Testing concept compositionality} 
\label{subsec:compositio,}

To test the compositionality of LLMs, we proceed in two stages: first, we measure the compositionality gap (\S\ref{subsub:gap_compos}), then we test the compositionality of the representations and the predictions (\S\ref{subsub:pred_compos}).

\subsubsection{Measuring the compositionality gap}
\label{subsub:gap_compos}
Instead of following \cite{PressZMSSL23} that measures the cases where a model correctly answers all constituent subquestions but fails on the corresponding compositional question to test multi-hop or relational chaining as \emph{compositional gap}, we probe LLMs through QA tasks with conjunctive concept compositions.
For each compositional question $q \in \mathcal{Q} $, let $\mathcal{S}(q)$ denote its set of sub-questions.
Let $\mathcal{Q}^c_t = \{q \in \mathcal{Q}_t   / SubAcc(q)=1\}$ be the set of compositional questions for a given task $t$ that has its subquestions correctly answered, where  $\mathrm{SubAcc}(q) = \mathbb{I}\!\Big[\forall s \in \mathcal{S}(q):\ \hat{y}(s) = y(s)\Big]$.\\
 
\noindent\textbf{Evaluation metrics}. To measure the compositionality gap, we define the following metrics:
\begin{itemize}[itemsep=0pt, topsep=0pt, leftmargin=10pt]
 
\item \textit{Per-question CGA.} We define the per-question compositional accuracy as:
\[
\mathrm{CGA}(q) =1-\mathrm{Acc}(q)=1 - \mathbb{I}\big[\hat{y}(q) = y(q) \big]\quad \forall q \in \mathcal{Q}^c_t
\]
 where $\bar{q}$ denote the negative version of question $\bar{q}$, $\mathcal{S}(q)$ denotes the set of subquestions of $q$, $\hat{y}(q)$ is the model prediction, and $y(q)$ is the ground truth answer.

\item\textit{Compositional Gap Accuracy (CGA).} The proportion of compositional questions answered incorrectly for which all corresponding subquestions were answered correctly:
\[
\mathrm{CGA}=\frac{1}{|\mathcal{Q}^c_t |}\sum_{q\in\mathcal{Q}^c_t }\mathrm{CGA}(q)
\]

\item \textit{Compositional Gap Consistency (CGC}). We define a consistency-based compositional gap using paired positive ($q$) and negative versions  ($\bar q$) of each question, which decouples true performance from random chance. CGC measures the proportion of compositional questions answered incorrectly for which all corresponding sub-questions where answered consistently correct:

\end{itemize}
\[
\mathrm{CGC}
=
\frac{1}{|\mathcal{Q}^c_t |}\sum_{q\in\mathcal{Q}^c_t }
\mathrm{CGA}(q)\cdot\mathrm{CGA}(\bar q)
\]

\noindent\textbf{Results.} The main results are reported in Figure~\ref{fig:merged_compos}. Unlike~\cite{PressZMSSL23}, in conjunctive-based compositionality, LLMs perform better on compositional questions than on atomic ones. Notably, for the Binary-QA task, \textit{Mistral-v0.3-7B} shows the smallest compositional gap in accuracy ($13\%$), suggesting relatively better compositional generalization. However, this does not translate into strong consistency: its rate is $17.1\%$, slightly above \textit{Mistral-v0.2-7B} ($15.9\%$), while both remain far from ideal. This advantage in gap size should be interpreted alongside absolute performance, which is higher for \textit{Qwen3-4B} ($59.7\%$). Unsurprisingly, smaller models show the largest compositional and consistency gaps. For the MCQ task, the llama3-8B model has the highest compositional accuracy ($54.5\%$) but with a higher compositional gap compared to small-size models. Interestingly, smaller models on this task show lower gaps than in the binary task, yet overall compositional accuracy remains very low ($<39\%$).
\begin{figure*}[ht]
    \centering
    \small
        \includegraphics[width=.8\textwidth]{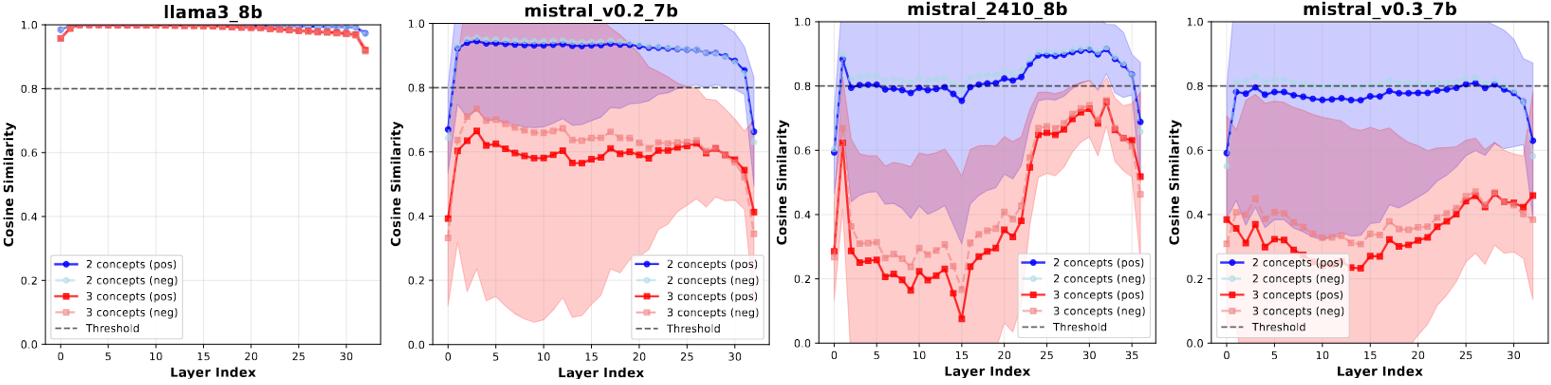}

    \caption{
    Binary-QA Task: Cosine similarity per layer between each compositional question embedding $q$ and its decomposition,
    considering 2-Concepts ($q_1+q_2$) and 3-Concepts ($q_1+q_2+q_3$). For MCQ task, we observe similar curves trend. Qwen models have very similar trend as llama3\_8b for both tasks.}
    \label{fig:compositionality_binary_vs_mcq}
\end{figure*}

\begin{figure*}[t]
    \centering
    \small
    \begin{subfigure}[t]{0.48\linewidth}
        \centering
        \includegraphics[width=\linewidth, trim=0cm 1.8cm 0cm 0cm, clip]{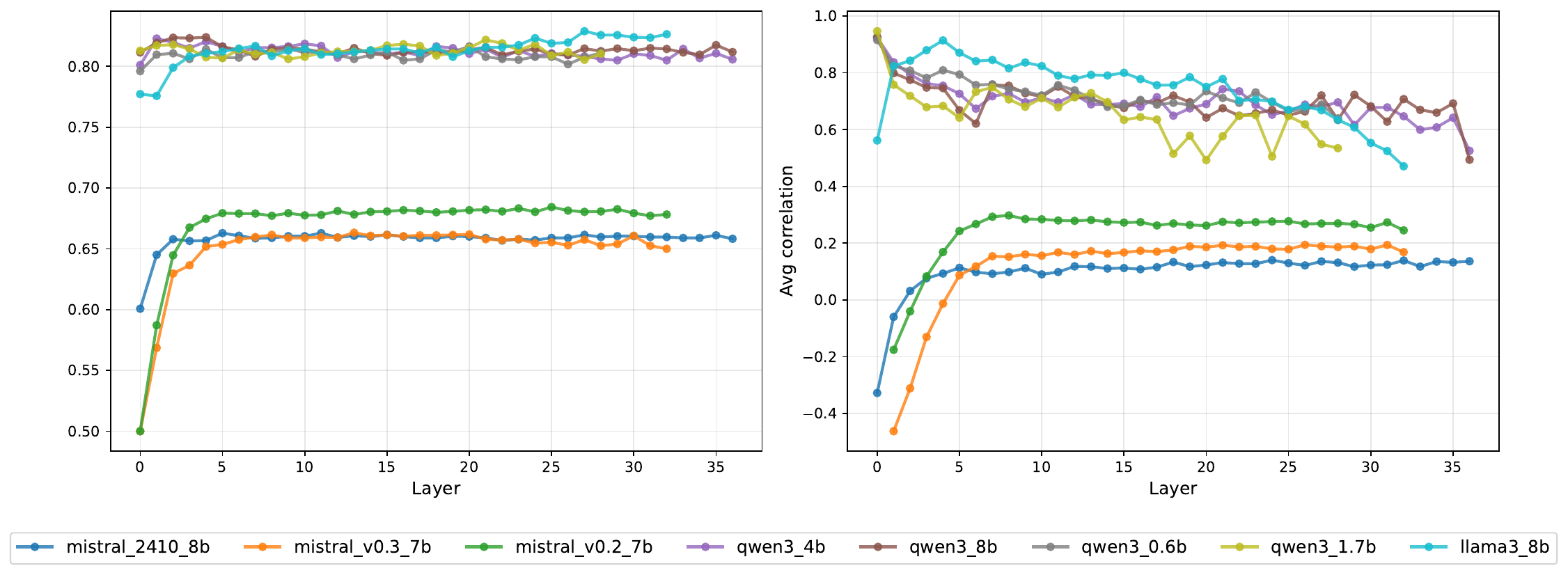}
        \caption{Binary Task. Mistral models lose logits correlation across layers, directly explaining performance gap compared to the other LLMs.}
        \label{fig:binary_task_corr}
    \end{subfigure}
    \hfill
    \begin{subfigure}[t]{0.48\linewidth}
        \centering
        \includegraphics[width=\linewidth, trim=0cm 1.8cm 0cm 0cm, clip]{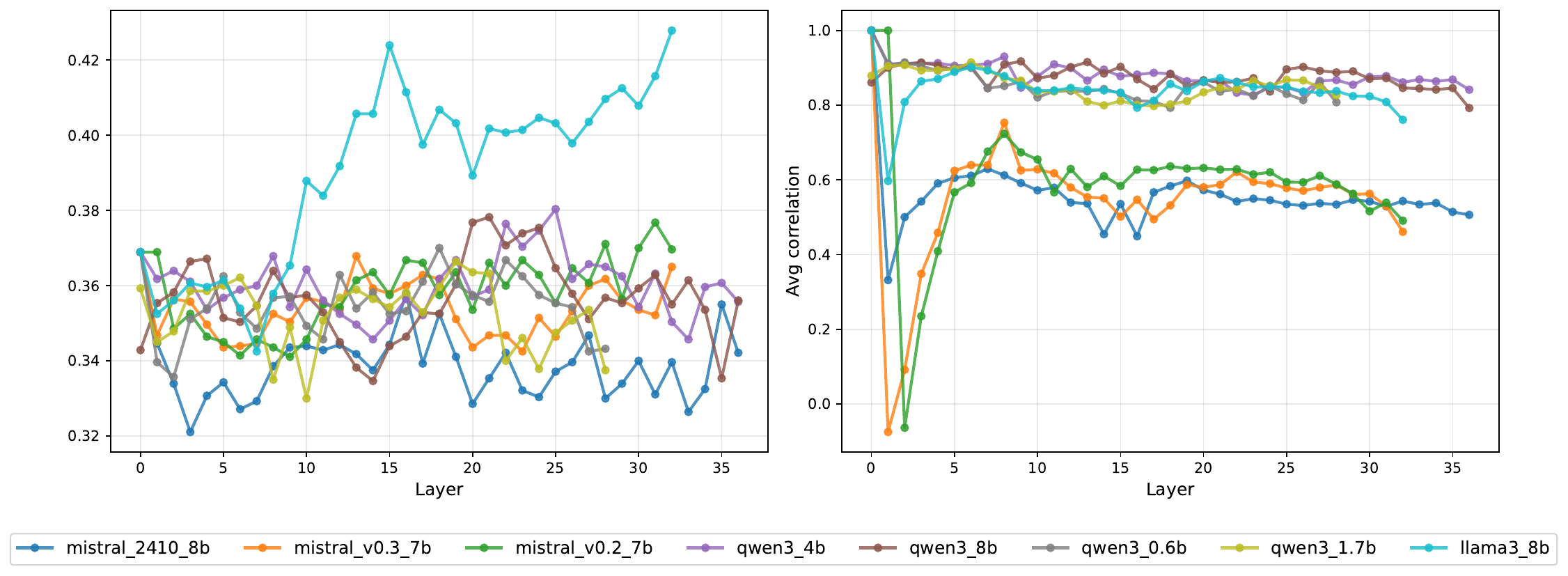}
        \caption{MCQ Task. Mistral models again exhibit weaker compositionality between atomics and composite questions compared to other models..}
        \label{fig:correlation_compos_mcq}
    \end{subfigure}

    \vspace{0.5em}

    \makebox[\textwidth][c]{%
        \includegraphics[width=0.8\textwidth]{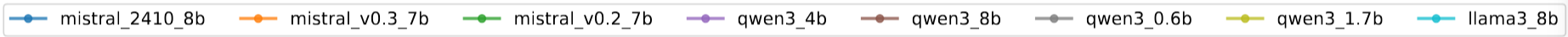}
    }
    \vspace{-0.6cm}
    \caption{Logits-based Pearson correlation coefficient across LLM layers for both tasks.}
    \label{fig:compositionality_across_tasks}
\end{figure*}

\subsubsection{Measuring compositionality}
\label{subsub:pred_compos}
Here, we attempt to answer two core questions: (i) Are concept representations compositional? (ii) Can LLMs' predictions be explained through compositional representations?\\

\noindent\textbf{Experimental design.} To investigate whether internal representations learned by selected LLMs exhibit compositional structure (i.e., question (i)), we  follow previous work \cite{Stein24}, by testing compositionality as approximate additivity. We analyze cosine similarity scores between $\mathbf{L}(q)$ and $\mathbf{L}_{\mathrm{add}}=\mathbf{L}(q_1) + \mathbf{L}(q_2)$, enabling systematic comparison of compositional behavior across models. We then compute a cosine similarity score $s\in[-1,1]$ as:
\[
s = \operatorname{sim}\big(\mathbf{L}(q),\,\mathbf{L}_{\mathrm{add}}\big)
= \frac{\mathbf{L}(q)^\top \mathbf{L}_{\mathrm{add}}}{\|\mathbf{L}(q)\|\,\|\mathbf{L}_{\mathrm{add}}\|}.
\]
The score $s$ measures how well the additive approximation by atomics matches the model's composite representation. 

To test prediction compositionality (i.e., question (ii)), we train multinomial logistic regression probes on frozen LLM layer representations rather than relying on the model’s native logits of the final layer only. This layer-wise probing framework enables analysis across all layers, revealing where compositional structure is encoded. By decoupling evaluation from LLMs' output-formatting artifacts of auto-regressive generation and focusing on fixed embeddings, the approach isolates representational content and yields a more robust assessment of how internal model features support compositional predictions. Probes are trained on a subset of factual items and evaluated on held-out facts, assessing not fact inference but the consistency of factual and compositional encoding within the model’s representational space and whether it emerges across LLMs from its frozen representation, and explain compositional prediction by its atomic predictions. 

To test the level of prediction compositionality, we measure correlations between predictions from the full composite representation \(\mathbf{L}(q)\) and the combination of atomic representations \(L_{\text{add}}\). This is performed using logits and embeddings, and probability-based methods. 
Let \(\mathbf{v}_{\text{comp}}\) denote the embedding of a composite question, \(\mathbf{v}_i\) the embeddings of its \(n\) atomic sub-questions, \(f\) the trained classifier, and \(\mathcal{L}(f(\mathbf{v}))\) the logit for the true answer class:

\begin{itemize}[itemsep=0pt, topsep=0pt, leftmargin=10pt]
    \item \textbf{Logit Additivity:} Correlation between $\mathcal{L}(f(\mathbf{v}_{\text{comp}}))$ and $\sum_{i=1}^{n} \mathcal{L}(f(\mathbf{v}_i))$, testing whether logit values compose linearly.
    
    \item \textbf{Embedding Summation:} Correlation between $P(f(\mathbf{v}_{\text{comp}}))$ and $P(f(\sum_{i=1}^{n} \mathbf{v}_i))$, evaluating whether summed embeddings predict the same class probabilities as composite embeddings.
    
    \item \textbf{Probability Averaging:} Correlation between $P(f(\mathbf{v}_{\text{comp}}))$ and $\frac{1}{n}\sum_{i=1}^{n} P(f(\mathbf{v}_i))$, testing whether individual atomic predictions aggregate to composite predictions.
\end{itemize}

Each metric probes a distinct compositionality hypothesis: decision-boundary linearity, embedding-space compositionality, and compositional confidence.

\noindent\textbf{Results.} From an \textbf{embedding-based perspective,} Figure~\ref{fig:compositionality_binary_vs_mcq} reports cosine similarity between compositional questions and their sub-questions across LLM layers. In both Binary-QA and MCQ tasks, Mistral models exhibit the lowest and most variable compositionality correlation. In both settings, 2-Concept compositions consistently show higher compositionality than 3-Concept ones. From a \textbf{prediction-based perspective}, Mistral achieves moderate Binary-QA accuracy ($\sim$65\%) with weak compositional correlations (logit correlations $\sim$0.4), and lower MCQ accuracy ($\sim$35\%). In contrast, Qwen and LLaMA attain higher Binary-QA accuracy ($\sim$80\%) and comparable MCQ performance ($\sim$35--39\%), while maintaining consistently strong correlations across settings (logits $\gtrsim$0.85, probabilities $\gtrsim$0.7), indicating a substantially stronger alignment between compositional structure and predictive behavior (cf. Table~\ref{tab:corr_combined_grouped}).

\begin{figure*}[t]
    \centering

    \begin{subfigure}[b]{0.5175\textwidth}
        \centering
        \includegraphics[width=\textwidth]{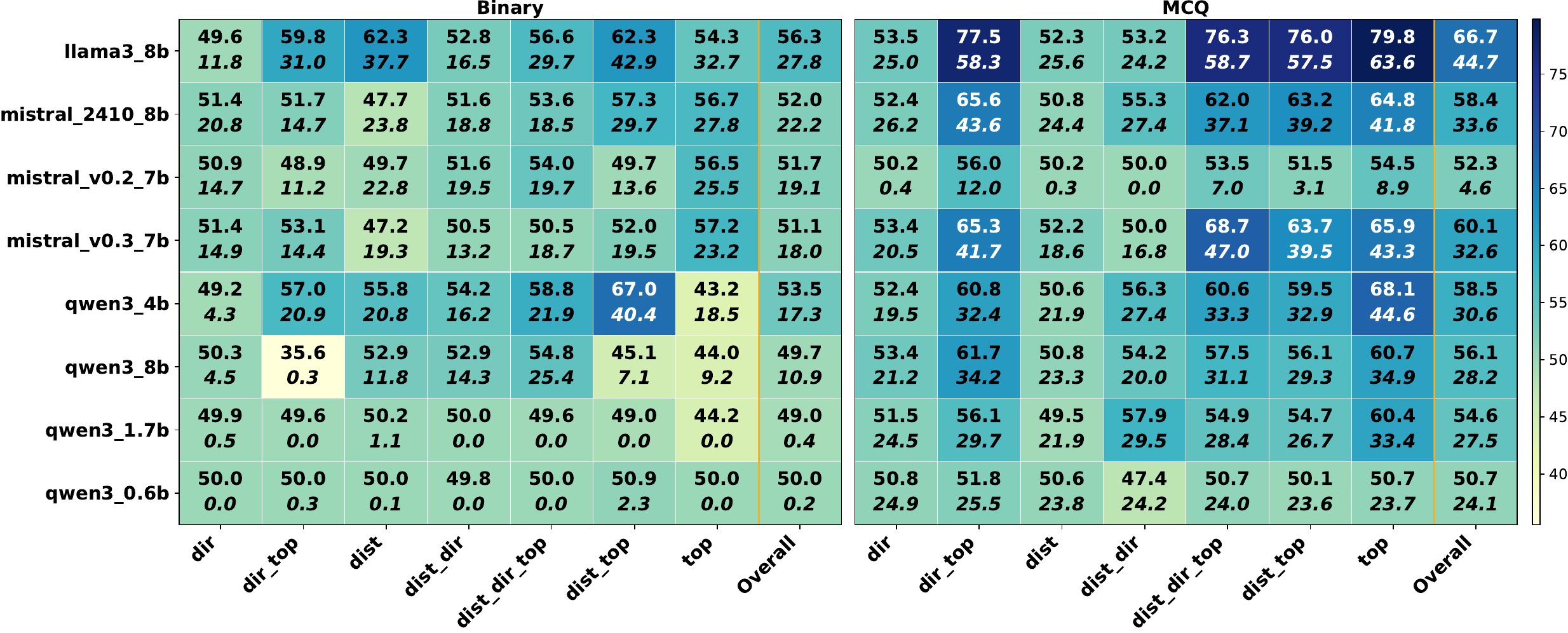}
        \caption{Per-concept and overall grounding performance.}
        \label{fig:grd_subfig_a}
    \end{subfigure}
    \hfill
    \begin{subfigure}[b]{0.4625\textwidth}
        \centering
        \includegraphics[width=\textwidth, trim=4.8cm 0cm 0cm 0cm, clip]{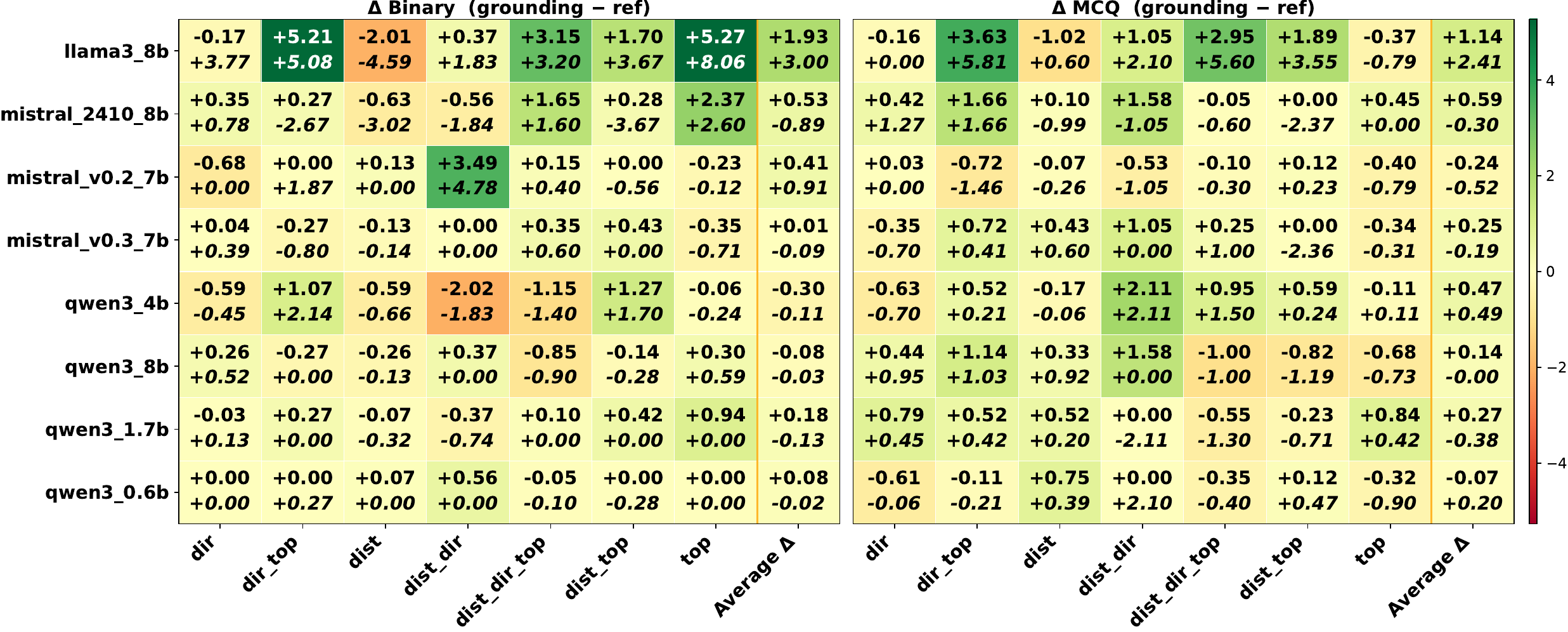}
        \caption{Performance changes in grounding (improvement/ degradation).}
        \label{fig:grd_subfig_b}
    \end{subfigure}
    \vspace{-0.3cm}
    \caption{Effect of grounding information on model performance compared to settings without grounding information per concept type. Consistency is reported below accuracy in each cell of the heatmaps, the final column shows the overall and average results.}
    \label{fig:grounding_context}
\end{figure*}

\noindent\textbf{To sum up}. Across experiments, LLMs exhibit a compositional consistency gap in QA tasks. In contrast, probing latent representations across layers reveals that predictive compositionality has a clear impact on performance: models with compositional embeddings preserve prediction consistency and achieve more robust results, unlike Mistral-like architectures.

\vspace{-0.2cm}
\definecolor{mycyan}{RGB}{0,160,160} 
\newcommand{\mcqcell}[1]{\cellcolor{mycyan!30}\textbf{#1}} 

\begin{table}[t]
    \centering
    \small
    \rowcolors{2}{white}{gray!7}
    \resizebox{\linewidth}{!}{%
    \begin{tabular}{l *{3}{cc} *{4}{cc} *{1}{cc}}
    \toprule
    & \multicolumn{6}{c}{\textbf{Mistral}} & \multicolumn{8}{c}{\textbf{Qwen}} & \multicolumn{2}{c}{\textbf{Llama}} \\
    \cmidrule(lr){2-7} \cmidrule(lr){8-15} \cmidrule(lr){16-17}
    Method
      & \multicolumn{2}{c}{7b\_v0.2} & \multicolumn{2}{c}{7b\_v0.3} & \multicolumn{2}{c}{8b\_2410}
      & \multicolumn{2}{c}{0.6b} & \multicolumn{2}{c}{1.7b} & \multicolumn{2}{c}{4b} & \multicolumn{2}{c}{8b}
      & \multicolumn{2}{c}{8b} \\
    \midrule
    \# layers
      & \multicolumn{2}{c}{33} & \multicolumn{2}{c}{33} & \multicolumn{2}{c}{37}
      & \multicolumn{2}{c}{29} & \multicolumn{2}{c}{29} & \multicolumn{2}{c}{37} & \multicolumn{2}{c}{37}
      & \multicolumn{2}{c}{33} \\ \midrule

    acc.(\%)
      & \mcqcell{67.03} & 35.84 & \mcqcell{64.88} & 35.39 & \mcqcell{65.75} & 33.89
      & \mcqcell{80.89} & 35.48 & \mcqcell{81.27} & 35.22 & \mcqcell{81.18} & 36.02
      & \mcqcell{81.36} & 35.66 & \mcqcell{81.32} & 39.01 \\ \midrule

    logits
      & \mcqcell{0.53} & 0.60 & \mcqcell{0.37} & 0.55 & \mcqcell{0.33} & 0.57
      & \mcqcell{0.98} & 0.86 & \mcqcell{0.99} & 0.86 & \mcqcell{0.99} & 0.88
      & \mcqcell{0.99} & 0.88 & \mcqcell{0.99} & 0.85 \\

    embed.
      & \mcqcell{0.23} & 0.51 & \mcqcell{0.05} & 0.44 & \mcqcell{0.08} & 0.45
      & \mcqcell{0.80} & 0.83 & \mcqcell{0.73} & 0.81 & \mcqcell{0.76} & 0.86
      & \mcqcell{0.78} & 0.84 & \mcqcell{0.58} & 0.72 \\

    proba.
      & \mcqcell{0.24} & 0.59 & \mcqcell{0.12} & 0.55 & \mcqcell{0.10} & 0.56
      & \mcqcell{0.73} & 0.85 & \mcqcell{0.66} & 0.85 & \mcqcell{0.70} & 0.88
      & \mcqcell{0.70} & 0.88 & \mcqcell{0.74} & 0.84 \\

    \bottomrule
    \end{tabular}%
    }
    \caption{Average correlations and accuracies for \textcolor{mycyan}{Binary}/MCQ.}
    \label{tab:corr_combined_grouped}
 \vspace{-1.15cm}
\end{table}
\subsection{Testing concept grounding}
\label{subsec:groundedness}
We test \emph{contextual grounding} via QA performance on tasks defined in a numerically explicit spatial context using coordinates, distances, and bearing angles, where all relevant real-world information is provided in context to LLMs as follows:
\begin{tcolorbox}[
  colback=cyan!10!white,
  colframe=cyan!70!black,
  boxsep=0pt,
  left=0pt, right=0pt, top=0pt, bottom=0pt,
  sharp corners,
  width=\columnwidth
]
\small
\noindent\textbf{\textcolor{cyan!70!black}{Context\_Binary:}} $X$ at $(a_X,b_X)$, $Y$ at $(a_Y,b_Y)$, distance $d(X,Y)=N$~km, bearing $\beta_{Y\to X}=\theta^\circ$, X and Y are considered close if $d(X,Y)\leq T_0$, and far otherwise.\\[2pt]
\noindent\textbf{\textcolor{cyan!70!black}{Context\_MCQ:}} $Y$ at $(a_Y,b_Y)$, $\{X_i\ \text{at}\ (a_i,b_i),\ d(Y,X_i)=N_i\ \text{km},\ \beta_{Y\to X_i}=\theta_i^\circ\}_{i=1}^{3}$ ($X_i$ ward option $i$), X and Y are considered close if $d(X,Y)\leq T_0$, and far otherwise.
\end{tcolorbox}

\textbf{Grounding.} In our context is defined by the ability to map linguistic concepts to their numerical meaning. If an LLM fully achieves grounding, it is expected to successfully answer the corresponding questions. In practice, answer quality provides a continuous measure of grounding. Direction- and distance-based concepts can be precisely tested in coordinate-based contexts, where all quantitative information is explicit. In contrast, topology-based concepts require a larger number of points, often leading to noisy contextual expansions and making practical evaluation challenging.

\noindent\textbf{Experimental design.} In this test, the LLM context is augmented with relevant geographical measurements and then asked to answer a corresponding related question. This protocol aims to evaluate LLMs' ability to ground concepts appearing in a given question to equivalent numerical representations.

\noindent\textbf{Results.} As shown in Figure~\ref{fig:grounding_context}, topology-independent spatial grounding performance in LLMs varies substantially across both task formulations and model architectures. In the MCQ setting, several models surpass the random baseline (33\%), with Qwen3-8B achieving the strongest results (66.7\% accuracy and 44.7\% consistency on the distance concept). However, overall consistency remains limited, indicating only weak grounding ability. In the Binary-QA setting, accuracy largely concentrates around chance level (\(\sim\)50\%) and is consistently paired with low consistency, revealing significant instability in concept grounding (e.g., Mistral-8B at \(\approx 52\%\)).
Across geographical concept-mapping tasks, LLMs demonstrate weak and inconsistent grounding, with no systematic advantage for direction- or distance-based concepts over topological ones. In several cases, performance degrades below that of simple factual recall, with only marginal improvement in some instances, achieving at best an average improvement of 1.93\% accuracy and 3\% consistency (Fig.~\ref{fig:grd_subfig_b}). 
 
\noindent \textbf{To sum up.} Concepts are not naturally grounded in the evaluated LLMs; instead, the results indicate a reliance on in-context concept memorization rather than true numerical meaning grounding. This aligns with prior findings showing that text-only small LMs struggle to map linguistic concepts to non-linguistic referents~\cite{patel2022mapping}. 
\section{Conclusion}

We investigated geo-spatial concept understanding in LLMs through the lens of abstraction, grounding, and compositionality. Our results show that while LLMs encode and partially compose concepts, factual consistency and real-world grounding remain critical bottlenecks.
The core of our probing concept-centric QA benchmark (i.e., questions and ground truth) is based on a task-agnostic generation and filtering of relational facts in the form of triplets (\S~ Algorithm \ref{alg:triplet_gen}, \ref{alg:gt_answers}). Such a widely adopted knowledge representation form inherently eases the extendability and reuse of our probing methodology to the study of abstraction, compositionality, and grounding of other concepts (e.g., truth \cite{azaria-mitchell-2023-internal}) even in other domains (e.g., gender bias in healthcare \cite{ahsan-etal-2025-elucidating}).\\
However, while providing valuable insights, our study has some limitations. First, our experiments rely only on two real-world geographic regions. Second, the studied concepts may not fully capture the  complexity of other real-world concepts. Third,  our probing experiments are limited to linear classifiers. \\
Our work has potential implications for information and knowledge management, including the following: (i) \textbf{Information access and retrieval}: our findings reveal that LLMs generally have a good level of performance in recognizing  OOD concept instances, indicating a significant ability of abstraction. Based on this finding,  new paths moving from the current designs of LLMs for relevance ranking \cite{pradeep2023rankvicuna,zhuang2024setwise,khramtsova2024leveraging} toward new approaches of  axiomatic IR \cite{Voske21} or mechanistic interpretability for IR \cite{Parry25}  are worth investigation by testing the concept of \textit{relevance} through its core properties that constrain words, documents, and queries; (ii) \textbf{Mining multimodal content}: the limited grounding of concepts in current LLMs provides concrete evidence of why  multi-modal (e.g. language and vision) models must go beyond traditional two-tower models \cite{RadfordKHRGASAM21}. Our findings argue for  externally grounded models that complement LLMs with parametric mechanisms for explicit grounding in the same line as parametric retrieval augmented generation \cite{Su25}; (iii) \textbf{Evaluation}: our findings reveal a critical limited ability of downstream task evaluation  to actually probe LLMs on their understanding of conceptual knowledge. This result calls for the design of new benchmarks suited to concept probing, annotated with core properties,  standardized metrics, and reference concepts, configurations with published baselines.\\

\bibliographystyle{ACM-Reference-Format}
\bibliography{general}

@STRING{jan = "Jan."}

@STRING{mar = "March"}

@STRING{jun = "June"}

@STRING{jul = "July"}

@STRING{aug = "Aug."}

@STRING{nov = "Nov."}

@STRING{dec = "Dec."}

@article{Goddu24,
title = {LLMs don't know anything: reply to Yildirim and Paul.},
journal = {Trends in Cognitive Sciences},
volume = {28},
number = {11},
pages = {963-964},
year = {2024},
issn = {1364-6613},
author = {Goddu, Mariel K. and Noë, Alva and Thompson, Evan},

}

@article{YILDIRIM2024404,
title = {From task structures to world models: what do LLMs know?},
journal = {Trends in Cognitive Sciences},
volume = {28},
number = {5},
pages = {404-415},
year = {2024},
issn = {1364-6613},
doi = {https://doi.org/10.1016/j.tics.2024.02.008},
url = {https://www.sciencedirect.com/science/article/pii/S1364661324000354},
author = {Ilker Yildirim and L.A. Paul},
}

@inproceedings{wang-etal-2024-knowledge-mechanisms,
    title = "Knowledge Mechanisms in Large Language Models: A Survey and Perspective",
    author = "Wang, Mengru  and
      Yao, Yunzhi  and
      Xu, Ziwen  and
      Qiao, Shuofei  and
      Deng, Shumin  and
      Wang, Peng  and
      Chen, Xiang  and
      Gu, Jia-Chen  and
      Jiang, Yong  and
      Xie, Pengjun  and
      Huang, Fei  and
      Chen, Huajun  and
      Zhang, Ningyu",
    editor = "Al-Onaizan, Yaser  and
      Bansal, Mohit  and
      Chen, Yun-Nung",
    booktitle = "Findings of the Association for Computational Linguistics: EMNLP 2024",
    month = nov,
    year = "2024",
    address = "Miami, Florida, USA",
    publisher = "Association for Computational Linguistics",
    url = "https://aclanthology.org/2024.findings-emnlp.416/",
    doi = "10.18653/v1/2024.findings-emnlp.416",
    pages = "7097--7135",
}

@inproceedings{koh2020concept,
  title={Concept bottleneck models},
  author={Koh, Pang Wei and Nguyen, Thao and Tang, Yew Siang and Mussmann, Stephen and Pierson, Emma and Kim, Been and Liang, Percy},
  booktitle={International conference on machine learning},
  pages={5338--5348},
  year={2020},
  organization={PMLR}
}

@article{Pavlick23,
    author = {Pavlick, Ellie},
    title = {Symbols and grounding in large language models},
    journal = {Philosophical Transactions of the Royal Society A: Mathematical, Physical and Engineering Sciences},
    volume = {381},
    number = {2251},
    pages = {20220041},
    year = {2023},
}

@article{belinkov-2022-probing,
    title = "Probing Classifiers: Promises, Shortcomings, and Advances",
    author = "Belinkov, Yonatan",
    journal = "Computational Linguistics",
    volume = "48",
    number = "1",
    month = mar,
    year = "2022",
    address = "Cambridge, MA",
    publisher = "MIT Press",
    url = "https://aclanthology.org/2022.cl-1.7/",
    doi = "10.1162/coli_a_00422",
    pages = "207--219"
}

@inproceedings{brachman-1979-taxonomy,
    title = "Taxonomy, Descriptions, and Individuals in Natural Language Understanding",
    author = "Brachman, Ronald J.",
    booktitle = "17th Annual Meeting of the Association for Computational Linguistics",
    month = jun,
    year = "1979",
    address = "La Jolla, California, USA",
    publisher = "Association for Computational Linguistics",
    url = "https://aclanthology.org/P79-1009/",
    doi = "10.3115/982163.982174",
    pages = "33--37"
}

@inproceedings{beinborn-etal-2018-multimodal,
    title = "Multimodal Grounding for Language Processing",
    author = "Beinborn, Lisa  and
      Botschen, Teresa  and
      Gurevych, Iryna",
    editor = "Bender, Emily M.  and
      Derczynski, Leon  and
      Isabelle, Pierre",
    booktitle = "Proceedings of the 27th International Conference on Computational Linguistics",
    month = aug,
    year = "2018",
    address = "Santa Fe, New Mexico, USA",
    publisher = "Association for Computational Linguistics",
    url = "https://aclanthology.org/C18-1197/",
    pages = "2325--2339"
}

@article{cohn2025evaluating,
  title={Evaluating the Ability of Large Language Models to Reason about Cardinal Directions, Revisited},
  author={Cohn, Anthony G and Blackwell, Robert E},
  journal={arXiv preprint arXiv:2507.12059},
  year={2025}
}

@article{van2025opportunities,
  title={Opportunities and challenges of integrating geographic information science and large language models},
  author={Van de Weghe, Nico and De Sloover, Lars and Cohn, Anthony and Huang, Haosheng and Scheider, Simon and Sieber, Ren{\'e}e and Timpf, Sabine and Claramunt, Christophe},
  journal={Journal of Spatial Information Science},
  number={30},
  pages={93--116},
  year={2025}
}

@article{zhang2025geoanalystbench,
  title={GeoAnalystBench: A GeoAI benchmark for assessing large language models for spatial analysis workflow and code generation},
  author={Zhang, Qianheng and Gao, Song and Wei, Chen and Zhao, Yibo and Nie, Ying and Chen, Ziru and Chen, Shijie and Su, Yu and Sun, Huan},
  journal={Transactions in GIS},
  volume={29},
  number={7},
  pages={e70135},
  year={2025},
  publisher={Wiley Online Library}
}

@article{ji2025foundation,
  title={Foundation models for geospatial reasoning: assessing the capabilities of large language models in understanding geometries and topological spatial relations},
  author={Ji, Yuhan and Gao, Song and Nie, Ying and Maji{\'c}, Ivan and Janowicz, Krzysztof},
  journal={International Journal of Geographical Information Science},
  pages={1--38},
  year={2025},
  publisher={Taylor \& Francis}
}

@article{Trager2023LinearSO,
  title={Linear Spaces of Meanings: Compositional Structures in Vision-Language Models},
  author={Matthew Trager and Pramuditha Perera and Luca Zancato and Alessandro Achille and Parminder Bhatia and Stefan 0 Soatto},
  journal={2023 IEEE/CVF International Conference on Computer Vision (ICCV)},
  year={2023},
  pages={15349-15358},
  url={https://api.semanticscholar.org/CorpusID:257766294}
}

@article{
yamada2024evaluating,
title={Evaluating Spatial Understanding of Large Language Models},
author={Yutaro Yamada and Yihan Bao and Andrew Kyle Lampinen and Jungo Kasai and Ilker Yildirim},
journal={Transactions on Machine Learning Research},
issn={2835-8856},
year={2024},
url={https://openreview.net/forum?id=xkiflfKCw3},
note={}
}

@inproceedings{
li2023can,
title={Can Language Models Understand Physical Concepts?},
author={Lei Li and Jingjing Xu and Qingxiu Dong and Ce Zheng and Xu Sun and Lingpeng Kong and Qi Liu},
booktitle={The 2023 Conference on Empirical Methods in Natural Language Processing},
year={2023},
url={https://openreview.net/forum?id=HaSS8a3Oe7}
}

@inproceedings{
park2025iclr,
title={{ICLR}: In-Context Learning of Representations},
author={Core Francisco Park and Andrew Lee and Ekdeep Singh Lubana and Yongyi Yang and Maya Okawa and Kento Nishi and Martin Wattenberg and Hidenori Tanaka},
booktitle={The Thirteenth International Conference on Learning Representations},
year={2025},
url={https://openreview.net/forum?id=pXlmOmlHJZ}
}

@article{yu2025understanding,
  title={Understanding and mitigating gender bias in llms via interpretable neuron editing},
  author={Yu, Zeping and Ananiadou, Sophia},
  journal={arXiv preprint arXiv:2501.14457},
  year={2025}
}

@inproceedings{lewis-etal-2024-clip,
    title = "Does {CLIP} Bind Concepts? Probing Compositionality in Large Image Models",
    author = "Lewis, Martha  and
      Nayak, Nihal  and
      Yu, Peilin  and
      Merullo, Jack  and
      Yu, Qinan  and
      Bach, Stephen  and
      Pavlick, Ellie",
    editor = "Graham, Yvette  and
      Purver, Matthew",
    booktitle = "Findings of the Association for Computational Linguistics: EACL 2024",
    month = mar,
    year = "2024",
    address = "St. Julian{'}s, Malta",
    publisher = "Association for Computational Linguistics",
    url = "https://aclanthology.org/2024.findings-eacl.101/",
    pages = "1487--1500"
}

@article{lovering-pavlick-2022-unit,
    title = "Unit Testing for Concepts in Neural Networks",
    author = "Lovering, Charles  and
      Pavlick, Ellie",
    editor = "Roark, Brian  and
      Nenkova, Ani",
    journal = "Transactions of the Association for Computational Linguistics",
    volume = "10",
    year = "2022",
    address = "Cambridge, MA",
    publisher = "MIT Press",
    url = "https://aclanthology.org/2022.tacl-1.69/",
    doi = "10.1162/tacl_a_00514",
    pages = "1193--1208",
}

@inproceedings{ahsan-etal-2025-elucidating,
    title = "Elucidating Mechanisms of Demographic Bias in {LLM}s for Healthcare",
    author = "Ahsan, Hiba  and
      Sen Sharma, Arnab  and
      Amir, Silvio  and
      Bau, David  and
      Wallace, Byron C",
    editor = "Christodoulopoulos, Christos  and
      Chakraborty, Tanmoy  and
      Rose, Carolyn  and
      Peng, Violet",
    booktitle = "Findings of the Association for Computational Linguistics: EMNLP 2025",
    month = nov,
    year = "2025",
    address = "Suzhou, China",
    publisher = "Association for Computational Linguistics",
    url = "https://aclanthology.org/2025.findings-emnlp.789/",
    doi = "10.18653/v1/2025.findings-emnlp.789",
    pages = "14614--14631",
    ISBN = "979-8-89176-335-7",
}

@inproceedings{azaria-mitchell-2023-internal,
    title = "The Internal State of an {LLM} Knows When It{'}s Lying",
    author = "Azaria, Amos  and
      Mitchell, Tom",
    editor = "Bouamor, Houda  and
      Pino, Juan  and
      Bali, Kalika",
    booktitle = "Findings of the Association for Computational Linguistics: EMNLP 2023",
    month = dec,
    year = "2023",
    address = "Singapore",
    publisher = "Association for Computational Linguistics",
    url = "https://aclanthology.org/2023.findings-emnlp.68/",
    doi = "10.18653/v1/2023.findings-emnlp.68",
    pages = "967--976",
}

@inproceedings{jin-etal-2025-exploring,
    title = "Exploring Concept Depth: How Large Language Models Acquire Knowledge and Concept at Different Layers?",
    author = "Jin, Mingyu  and
      Yu, Qinkai  and
      Huang, Jingyuan  and
      Zeng, Qingcheng  and
      Wang, Zhenting  and
      Hua, Wenyue  and
      Zhao, Haiyan  and
      Mei, Kai  and
      Meng, Yanda  and
      Ding, Kaize  and
      Yang, Fan  and
      Du, Mengnan  and
      Zhang, Yongfeng",
    editor = "Rambow, Owen  and
      Wanner, Leo  and
      Apidianaki, Marianna  and
      Al-Khalifa, Hend  and
      Eugenio, Barbara Di  and
      Schockaert, Steven",
    booktitle = "Proceedings of the 31st International Conference on Computational Linguistics",
    month = jan,
    year = "2025",
    address = "Abu Dhabi, UAE",
    publisher = "Association for Computational Linguistics",
    url = "https://aclanthology.org/2025.coling-main.37/",
    pages = "558--573",
}

@inproceedings{geva-etal-2022-transformer,
    title = "Transformer Feed-Forward Layers Build Predictions by Promoting Concepts in the Vocabulary Space",
    author = "Geva, Mor  and
      Caciularu, Avi  and
      Wang, Kevin  and
      Goldberg, Yoav",
    editor = "Goldberg, Yoav  and
      Kozareva, Zornitsa  and
      Zhang, Yue",
    booktitle = "Proceedings of the 2022 Conference on Empirical Methods in Natural Language Processing",
    month = dec,
    year = "2022",
    address = "Abu Dhabi, United Arab Emirates",
    publisher = "Association for Computational Linguistics",
    url = "https://aclanthology.org/2022.emnlp-main.3/",
    doi = "10.18653/v1/2022.emnlp-main.3",
    pages = "30--45",
}

@article{Nainani2024EvaluatingBM,
  title={Evaluating Brain-Inspired Modular Training in Automated Circuit Discovery for Mechanistic Interpretability},
  author={Jatin Nainani},
  journal={ArXiv},
  year={2024},
  volume={abs/2401.03646},
  url={https://api.semanticscholar.org/CorpusID:266843932}
}

@article{Fong2018Net2VecQA,
  title={Net2Vec: Quantifying and Explaining How Concepts are Encoded by Filters in Deep Neural Networks},
  author={Ruth Fong and Andrea Vedaldi},
  journal={2018 IEEE/CVF Conference on Computer Vision and Pattern Recognition},
  year={2018},
  pages={8730-8738},
  url={https://api.semanticscholar.org/CorpusID:2738204}
}

@article{Poeta25,
author = {Poeta, Eleonora and Ciravegna, Gabriele and Pastor, Eliana and Cerquitelli, Tania and Baralis, Elena},
title = {Concept-based Explainable Artificial Intelligence: A Survey},
year = {2025},
publisher = {Association for Computing Machinery},
address = {New York, NY, USA},
issn = {0360-0300},
url = {https://doi.org/10.1145/3774643},
doi = {10.1145/3774643},
note = {Just Accepted},
journal = {ACM Comput. Surv.},
month = nov
}

@article{schwalbe2022concept,
  title={Concept embedding analysis: A review},
  author={Schwalbe, Gesina},
  journal={arXiv preprint arXiv:2203.13909},
  year={2022}
}

@InProceedings{Goguen2005,
author="Goguen, Joseph",
editor="Dau, Frithjof
and Mugnier, Marie-Laure
and Stumme, Gerd",
title="What Is a Concept?",
booktitle="Conceptual Structures: Common Semantics for Sharing Knowledge",
year="2005",
publisher="Springer Berlin Heidelberg",
address="Berlin, Heidelberg",
pages="52--77",
isbn="978-3-540-31885-9"
}

@inproceedings{
dihan2025mapeval,
title={MapEval: A Map-Based Evaluation of Geo-Spatial Reasoning in Foundation Models},
author={Mahir Labib Dihan and MD Tanvir Hassan and MD TANVIR PARVEZ and Md Hasebul Hasan and Md Almash Alam and Muhammad Aamir Cheema and Mohammed Eunus Ali and Md Rizwan Parvez},
booktitle={Forty-second International Conference on Machine Learning},
year={2025},
url={https://openreview.net/forum?id=hS2Ed5XYRq}
}

@inproceedings{MaiJ0CL21,
  author={Gengchen Mai and Krzysztof Janowicz and Rui Zhu and Ling Cai and Ni Lao},
  title={Geographic Question Answering: Challenges, Uniqueness, Classification, and Future Directions},
  year={2021},
  cdate={1609459200000},
  pages={8},
  url={https://doi.org/10.5194/agile-giss-2-8-2021},
  booktitle={AGILE Conf.},
}

@inproceedings{ManviIcml24,
author = {Manvi, Rohin and Khanna, Samar and Burke, Marshall and Lobell, David and Ermon, Stefano},
title = {Large language models are geographically biased},
year = {2024},
publisher = {JMLR.org},
booktitle = {Proceedings of the 41st International Conference on Machine Learning},
articleno = {1409},
numpages = {16},
location = {Vienna, Austria},
series = {ICML'24}
}

@article{Mai24,
author = {Mai, Gengchen and Huang, Weiming and Sun, Jin and Song, Suhang and Mishra, Deepak and Liu, Ninghao and Gao, Song and Liu, Tianming and Cong, Gao and Hu, Yingjie and Cundy, Chris and Li, Ziyuan and Zhu, Rui and Lao, Ni},
title = {On the Opportunities and Challenges of Foundation Models for GeoAI (Vision Paper)},
year = {2024},
issue_date = {June 2024},
publisher = {Association for Computing Machinery},
address = {New York, NY, USA},
volume = {10},
number = {2},
issn = {2374-0353},
url = {https://doi.org/10.1145/3653070},
doi = {10.1145/3653070},
journal = {ACM Trans. Spatial Algorithms Syst.},
month = jul,
articleno = {11},
numpages = {46}
}

@article{Ramrakhiyaniipm2024,
author = {Ramrakhiyani, Nitin and Varma, Vasudeva and Palshikar, Girish Keshav and Pawar, Sachin},
title = {Gauging, enriching and applying geography knowledge in Pre-trained Language Models},
year = {2025},
issue_date = {Jan 2025},
publisher = {Pergamon Press, Inc.},
address = {USA},
volume = {62},
number = {1},
issn = {0306-4573},
url = {https://doi.org/10.1016/j.ipm.2024.103892},
doi = {10.1016/j.ipm.2024.103892},
journal = {Inf. Process. Manage.},
month = jan,
numpages = {23}
}

@inproceedings{
patel2022mapping,
title={Mapping Language Models to Grounded Conceptual Spaces},
author={Roma Patel and Ellie Pavlick},
booktitle={International Conference on Learning Representations},
year={2022},
url={https://openreview.net/forum?id=gJcEM8sxHK}
}

@inproceedings{
manvi2024geollm,
title={Geo{LLM}: Extracting Geospatial Knowledge from Large Language Models},
author={Rohin Manvi and Samar Khanna and Gengchen Mai and Marshall Burke and David B. Lobell and Stefano Ermon},
booktitle={The Twelfth International Conference on Learning Representations},
year={2024},
url={https://openreview.net/forum?id=TqL2xBwXP3}
}

@inproceedings{
gurnee2024language,
title={Language Models Represent Space and Time},
author={Wes Gurnee and Max Tegmark},
booktitle={The Twelfth International Conference on Learning Representations},
year={2024},
url={https://openreview.net/forum?id=jE8xbmvFin}
}

@journal{Fodor1975-FODTLO,
	title = {Connectionism and cognitive architecture: A critical analysis},
journal = {Cognition},
volume = {28},
number = {1},
pages = {3-71},
year = {1988},
issn = {0010-0277},
doi = {https://doi.org/10.1016/0010-0277(88)90031-5},
url = {https://www.sciencedirect.com/science/article/pii/0010027788900315},
author = {Jerry A. Fodor and Zenon W. Pylyshyn},
}

@inproceedings{PressZMSSL23,
  author       = {Ofir Press and
                  Muru Zhang and
                  Sewon Min and
                  Ludwig Schmidt and
                  Noah A. Smith and
                  Mike Lewis},
  editor       = {Houda Bouamor and
                  Juan Pino and
                  Kalika Bali},
  title        = {Measuring and Narrowing the Compositionality Gap in Language Models},
  booktitle    = {Findings of the Association for Computational Linguistics: {EMNLP}
                  2023, Singapore, December 6-10, 2023},
  pages        = {5687--5711},
  publisher    = {Association for Computational Linguistics},
  year         = {2023},
  url          = {https://doi.org/10.18653/v1/2023.findings-emnlp.378},
  doi          = {10.18653/V1/2023.FINDINGS-EMNLP.378},
  bibsource    = {dblp computer science bibliography, https://dblp.org}
}

@inproceedings{Stein24,
author = {Stein, Adam and Naik, Aaditya and Wu, Yinjun and Naik, Mayur and Wong, Eric},
title = {Towards compositionality in concept learning},
year = {2024},
publisher = {JMLR.org},
booktitle = {Proceedings of the 41st International Conference on Machine Learning},
articleno = {1893},
numpages = {26},
location = {Vienna, Austria},
series = {ICML'24}
}

@book{rosen2011discrete,
  author    = {Kenneth H. Rosen},
  title     = {Discrete Mathematics and Its Applications},
  edition   = {7},
  publisher = {McGraw--Hill Education},
  year      = {2011}
}

@inproceedings{mikolov2013distributed,
  title     = {Distributed Representations of Words and Phrases and their Compositionality},
  author    = {Tomas Mikolov and Ilya Sutskever and Kai Chen and Greg Corrado and Jeffrey Dean},
  booktitle = {Advances in Neural Information Processing Systems},
  year      = {2013},
  pages     = {3111--3119},
  note      = {NeurIPS 2013}
}

@article{naito2021revisiting,
  title   = {Revisiting Additive Compositionality: AND, OR and NOT Operations with Word Embeddings},
  author  = {Masahiro Naito and Sho Yokoi and Geewook Kim and Hidetoshi Shimodaira},
  journal = {Proceedings of the ACL-IJCNLP 2021 Student Research Workshop},
  year    = {2021}
}

@inproceedings{lake2018generalization,
  title     = {Generalization without Systematicity: On the Compositional Skills of Sequence-to-Sequence Recurrent Networks},
  author    = {Brenden M. Lake and Marco Baroni},
  booktitle = {Proceedings of the 35th International Conference on Machine Learning (ICML)},
  year      = {2018},
  pages     = {2873--2882},
  note      = {Proc. of ML Research Vol. 80}
}

@inproceedings{RadfordKHRGASAM21,
  author       = {Alec Radford and
                  Jong Wook Kim and
                  Chris Hallacy and
                  Aditya Ramesh and
                  Gabriel Goh and
                  Sandhini Agarwal and
                  Girish Sastry and
                  Amanda Askell and
                  Pamela Mishkin and
                  Jack Clark and
                  Gretchen Krueger and
                  Ilya Sutskever},
  editor       = {Marina Meila and
                  Tong Zhang},
  title        = {Learning Transferable Visual Models From Natural Language Supervision},
  booktitle    = {Proceedings of the 38th International Conference on Machine Learning,
                  {ICML} 2021, 18-24 July 2021, Virtual Event},
  series       = {Proceedings of Machine Learning Research},
  volume       = {139},
  pages        = {8748--8763},
  publisher    = {{PMLR}},
  year         = {2021},
  url          = {http://proceedings.mlr.press/v139/radford21a.html},
  bibsource    = {dblp computer science bibliography, https://dblp.org}
}

@inproceedings{Dumitru25,
  author       = {Alexandru Dumitru and
                  Venktesh V and
                  Adam Jatowt and
                  Avishek Anand},
  editor       = {Hamed Zamani and
                  Laura Dietz and
                  Benjamin Piwowarski and
                  Sebastian Bruch},
  title        = {Evaluating List Construction and Temporal Understanding capabilities
                  of Large Language Models},
  booktitle    = {Proceedings of the 2025 International {ACM} {SIGIR} Conference on
                  Innovative Concepts and Theories in Information Retrieval, {ICTIR}
                  2025, Padua, Italy, 18 July 2025},
  pages        = {369--379},
  publisher    = {{ACM}},
  year         = {2025},
  url          = {https://doi.org/10.1145/3731120.3744606},
  doi          = {10.1145/3731120.3744606},
  bibsource    = {dblp computer science bibliography, https://dblp.org}
}

@inproceedings{Voske21,
author = {V\"{o}lske, Michael and Bondarenko, Alexander and Fr\"{o}be, Maik and Stein, Benno and Singh, Jaspreet and Hagen, Matthias and Anand, Avishek},
title = {Towards Axiomatic Explanations for Neural Ranking Models},
year = {2021},
isbn = {9781450386111},
publisher = {Association for Computing Machinery},
address = {New York, NY, USA},
url = {https://doi.org/10.1145/3471158.3472256},
doi = {10.1145/3471158.3472256},
pages = {13–22},
numpages = {10},
location = {Virtual Event, Canada},
series = {ICTIR '21}
}

@inproceedings{Su25,
author = {Su, Weihang and Tang, Yichen and Ai, Qingyao and Yan, Junxi and Wang, Changyue and Wang, Hongning and Ye, Ziyi and Zhou, Yujia and Liu, Yiqun},
title = {Parametric Retrieval Augmented Generation},
year = {2025},
isbn = {9798400715921},
publisher = {Association for Computing Machinery},
address = {New York, NY, USA},
url = {https://doi.org/10.1145/3726302.3729957},
doi = {10.1145/3726302.3729957}
}

@inproceedings{Xie25,
author = {Xie, Yuzhang and Lu, Jiaying and Ho, Joyce and Nahab, Fadi and Hu, Xiao and Yang, Carl},
title = {PromptLink: Leveraging Large Language Models for Cross-Source Biomedical Concept Linking},
year = {2024},
isbn = {9798400704314},
publisher = {Association for Computing Machinery},
address = {New York, NY, USA},
url = {https://doi.org/10.1145/3626772.3657904},
doi = {10.1145/3626772.3657904},

booktitle = {Proceedings of the 47th International ACM SIGIR Conference on Research and Development in Information Retrieval},
pages = {2589–2593},
numpages = {5},
location = {Washington DC, USA},
series = {SIGIR '24}
}

@inproceedings{Parry25,
author = {Parry, Andrew and Chen, Catherine and Eickhoff, Carsten and MacAvaney, Sean},
title = {MechIR: A Mechanistic Interpretability Framework for Information Retrieval},
year = {2025},
isbn = {978-3-031-88719-2},
publisher = {Springer-Verlag},
address = {Berlin, Heidelberg},
url = {https://doi.org/10.1007/978-3-031-88720-8_16},
doi = {10.1007/978-3-031-88720-8_16},
booktitle = {Advances in Information Retrieval: 47th European Conference on Information Retrieval, ECIR 2025, Lucca, Italy, April 6–10, 2025, Proceedings, Part V},
pages = {89–95},
numpages = {7},
location = {Lucca, Italy}
}

@inproceedings{polley2022xvision,
  author    = {Sayantan Polley and Subhajit Mondal and Venkata Srinath Mannam and Kushagra Kumar and Subhankar Patra and Andreas N{\"u}rnberger},
  title     = {X-Vision: Explainable Image Retrieval by Re-Ranking in Semantic Space},
  booktitle = {Proceedings of the 31st ACM International Conference on Information \& Knowledge Management (CIKM '22)},
  pages     = {4955--4959},
  year      = {2022},
  doi       = {10.1145/3511808.3557187},
  url       = {https://doi.org/10.1145/3511808.3557187}
}

@inproceedings{ji2023ccr,
  author    = {Jianchao Ji and Zelong Li and Shuyuan Xu and Max Xiong and Juntao Tan and Yingqiang Ge and Hao Wang and Yongfeng Zhang},
  title     = {Counterfactual Collaborative Reasoning},
  booktitle = {Proceedings of the 16th ACM International Conference on Web Search and Data Mining (WSDM '23)},
  pages     = {249--257},
  year      = {2023},
  doi       = {10.1145/3539597.3570464},
  url       = {https://doi.org/10.1145/3539597.3570464}
}

@inproceedings{deng2024k2,
  author    = {Cheng Deng and Tianhang Zhang and Zhongmou He and Yi Xu and Qiyuan Chen and Yuanyuan Shi and Luoyi Fu and Weinan Zhang and Xinbing Wang and Chenghu Zhou and Zhouhan Lin and Junxian He},
  title     = {K2: A Foundation Language Model for Geoscience Knowledge Understanding and Utilization},
  booktitle = {Proceedings of the 17th ACM International Conference on Web Search and Data Mining (WSDM '24)},
  pages     = {161--170},
  year      = {2024},
  doi       = {10.1145/3616855.3635772},
  url       = {https://doi.org/10.1145/3616855.3635772}
}

@inproceedings{khramtsova2024leveraging,
  title={Leveraging llms for unsupervised dense retriever ranking},
  author={Khramtsova, Ekaterina and Zhuang, Shengyao and Baktashmotlagh, Mahsa and Zuccon, Guido},
  booktitle={Proceedings of the 47th International ACM SIGIR Conference on Research and Development in Information Retrieval},
  pages={1307--1317},
  year={2024}
}

@article{pradeep2023rankvicuna,
  title={Rankvicuna: Zero-shot listwise document reranking with open-source large language models},
  author={Pradeep, Ronak and Sharifymoghaddam, Sahel and Lin, Jimmy},
  journal={arXiv preprint arXiv:2309.15088},
  year={2023}
}

@inproceedings{zhuang2024setwise,
  title={A setwise approach for effective and highly efficient zero-shot ranking with large language models},
  author={Zhuang, Shengyao and Zhuang, Honglei and Koopman, Bevan and Zuccon, Guido},
  booktitle={Proceedings of the 47th International ACM SIGIR Conference on Research and Development in Information Retrieval},
  pages={38--47},
  year={2024}
}

@String{Computing = "Computing" }

@String{Computer = "{IEEE} Computer" }

@String{Springer = "Springer-Verlag" }

@BOOK{test,
   author = "Donald E. Knuth",
   title = "Seminumerical Algorithms",
   volume = 2,
   series = "The Art of Computer Programming",
   publisher = "Addison-Wesley",
   address = "Reading, MA",
   edition = "2nd",
   month = "10~" # jan,
   year = "1981",
}

@ArtifactSoftware{R,
    title = {R: A Language and Environment for Statistical Computing},
    author = {{R Core Team}},
    organization = {R Foundation for Statistical Computing},
    address = {Vienna, Austria},
    year = {2019},
    url = {https://www.R-project.org/},
}
\end{sloppypar}
\end{document}